\documentclass[letterpaper]{article} 
\usepackage{aaai2026}  
\usepackage{times}  
\usepackage{helvet}  
\usepackage{courier}  
\usepackage[hyphens]{url}  
\usepackage{graphicx} 
\usepackage{natbib}  
\usepackage{caption} 
\usepackage{algorithm}
\usepackage{algorithmic}

\usepackage{multirow}
\usepackage{array}
\usepackage{amsmath}
\usepackage{amsfonts}
\usepackage{amssymb}
\usepackage{makecell}
\usepackage{booktabs}
\usepackage{bbding}
\usepackage{multirow}
\usepackage{array}
\usepackage{amsmath}
\usepackage{amsfonts}
\usepackage{amssymb}
\usepackage{makecell}
\usepackage{booktabs}
\usepackage{bbding}
\usepackage[figuresright]{rotating}
\usepackage{booktabs}   
\usepackage{makecell}   
\usepackage{array}      

\usepackage{pgfplots}
\usepackage{xcolor}
\definecolor{barcolor}{HTML}{E07B67}

\usepackage{newfloat}
\usepackage{listings}
\DeclareCaptionStyle{ruled}{labelfont=normalfont,labelsep=colon,strut=off} 
\floatstyle{ruled}
\newfloat{listing}{tb}{lst}{}
\floatname{listing}{Listing}
\title{AlignSurvey: A Comprehensive Benchmark for Human Preferences Alignment in Social Surveys}

\title{AlignSurvey: A Comprehensive Benchmark for Human Preferences Alignment in Social Surveys}
\author {
    Chenxi Lin\textsuperscript{\rm 1},
    Weikang Yuan\textsuperscript{\rm 1},
    Zhuoren Jiang\textsuperscript{\rm 1,3}\thanks{Corresponding author},
    Biao Huang\textsuperscript{\rm 1,3}, \\
    Ruitao Zhang\textsuperscript{\rm 1},
    Jianan Ge\textsuperscript{\rm 1},
    Yueqian Xu\textsuperscript{\rm 2,3},
    Jianxing Yu\textsuperscript{\rm 2,3}
}
\affiliations {
    \textsuperscript{\rm 1}Zhejiang University\\
    \textsuperscript{\rm 2}Zhejiang Gongshang University\\
    \textsuperscript{\rm 3}Laboratory for Statistical Monitoring and Intelligent Governance of Common Prosperity, Zhejiang Gongshang University \\
    linchenxi@zju.edu.cn, 
    yuanwk@zju.edu.cn, 
    jiangzhuoren@zju.edu.cn,
    biaohuang@zju.edu.cn, \\
    zhangruitao24@zju.edu.cn,
    iamjianange@zju.edu.cn,
    xuyueqian@zjgsu.edu.cn,
    yujianxing@zju.edu.cn,
    
}

\usepackage{bibentry}

\newcommand{\percbar}[1]{%
    \begin{tikzpicture}
        \fill [barcolor] (0,0) rectangle (#1/100*3,0.18);
        \draw [lightgray] (0,0) rectangle (3,0.18);
    \end{tikzpicture}%
}

\begin{document}

\maketitle

\begin{abstract}
Understanding human attitudes, preferences, and behaviors through social surveys is essential for academic research and policymaking. Yet traditional surveys face persistent challenges, including fixed-question formats, high costs, limited adaptability, and difficulties ensuring cross-cultural equivalence. While recent studies explore large language models (LLMs) to simulate survey responses, most are limited to structured questions, overlook the entire survey process, and risks under-representing marginalized groups due to training data biases.
We introduce \textbf{AlignSurvey}, the first benchmark that systematically replicates and evaluates the full social survey pipeline using LLMs. It defines four tasks aligned with key survey stages: social role modeling, semi-structured interview modeling, attitude stance modeling and survey response modeling. It also provides task-specific evaluation metrics to assess alignment fidelity, consistency, and fairness at both individual and group levels, with a focus on demographic diversity.
To support AlignSurvey, we construct a multi-tiered dataset architecture: (i) the Social Foundation Corpus, a cross-national resource with 44K+ interview dialogues and 400K+ structured survey records; and (ii) a suite of Entire-Pipeline Survey Datasets, including the expert-annotated AlignSurvey-Expert (ASE) and two nationally representative surveys for cross-cultural evaluation.
We release the SurveyLM family, obtained through two-stage fine-tuning of open-source LLMs, and offer reference models for evaluating domain-specific alignment. All datasets, models, and tools are available at github and huggingface to support transparent and socially responsible research.

\end{abstract}

\begin{links}
    \link{Code}{https://github.com/PiLab-ZJU/AlignSurvey}
    \link{Datasets \& Models}{https://huggingface.co/PiLabZJU}
    \link{Extended Version}{https://arxiv.org/abs/2511.07871}
\end{links}

\section{Introduction}

Understanding human preferences, attitudes, and behaviors is central to both academic research and evidence-based policymaking. Social surveys, spanning qualitative interviews and quantitative questionnaires, have long served as a critical tool in this endeavor~\cite{wright2010survey,tourangeau2004survey}, with an estimated \$35.1 billion spent annually on survey research worldwide~\cite{esomar2024global}.

Despite this scale, traditional surveys face persistent challenges: fixed-question formats limit adaptability~\cite{heffetz2019difficulty}; high costs often force sampling compromises, introducing biases~\cite{kalton2009methods}; slow turnaround hinders responsiveness to emerging issues~\cite{moy2016problems,evans2018value,prosser2018twilight}; and cross-cultural equivalence remains difficult despite translation efforts~\cite{tsai2025challenges}.

Large language models (LLMs) offer a promising alternative. By learning from vast public corpora, LLMs can simulate human responses and reduce the burden of manual data collection~\cite{thapa2025large,mellon2024ais,zhang2025socioverse}. However, their outputs often reflect the preferences of digitally active, well-educated users~\cite{giorgi2025human,abeliuk2025fairness}, reinforcing representational bias~\cite{wang2025large,hu2025generative,hofmann2024ai} and marginalizing rural, low-income, or elderly populations. Group-level disparities are often underexplored, and many existing works lack systematic evaluation across demographic subgroups.

\begin{figure*}[t]
  \centering
  \includegraphics[width=\linewidth]{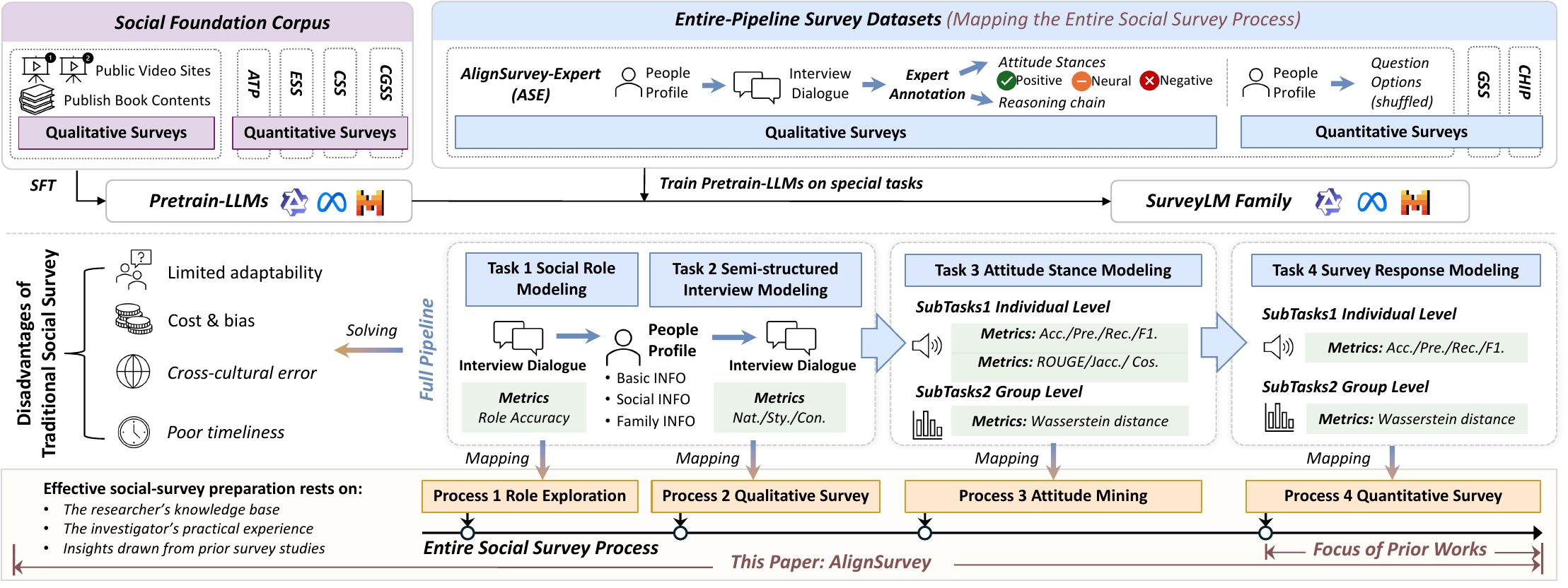}
  \caption{Overview of the \textbf{AlignSurvey}. AlignSurvey is a four-stage benchmark that mirrors the professional social survey process.
  The upper panel depicts a multi-tiered dataset: we pretrain on the Social Foundation Corpus for broad social science knowledge and fine-tune on the Entire-Pipeline Survey Dataset for the four survey stages: \emph{(1) Social Role Modeling; (2) Semi‑Structured Interview Modeling; (3) Attitude Stance Modeling; (4) Structured Response Modeling}. AlignSurvey is the first to align LLMs across the entire social science survey, surpassing prior work limited to structured responses.}
  \label{fig:overview}
\end{figure*}

Existing benchmarks primarily focus on fixed-option quantitative tasks~\cite{ji2023beavertails,liu2025towards,zhou2025fair,lee2024can}, overlooking the full pipelines of professional surveys, such as qualitative interviewing and context-aware reasoning. This critical gap calls for a comprehensive framework that aligns LLMs with human preferences across the entire survey process.

To address these gaps, we introduce \textbf{AlignSurvey}, the first benchmark designed to systematically replicate and evaluate the \textbf{full pipeline} of professional social surveys using LLMs. 
AlignSurvey mirrors four core stages of professional social surveys \cite{ahmed2024mixed,fetters2013achieving} by mapping role exploration, qualitative survey, attitude mining and quantitative survey to corresponding modeling tasks: \textit{Social Role Modeling}, \textit{Semi-structured Interview Modeling}, \textit{Attitude Stance Modeling} and \textit{Structured Response Modeling}. We introduce task-specific evaluation metrics that enable alignment assessment across tasks at individual and group levels, with a particular focus on fairness and demographic diversity.

To support this pipeline, we construct a multi-tiered dataset architecture. The first component, the \textbf{Social Foundation Corpus}, is a cross-national resource comprising 44,000+ interview dialogues collected from publicly accessible video platforms and oral history books, and 400,000+ structured records from four authoritative surveys: ATP, ESS, CSS, and CGSS. These corpora provide foundational knowledge across diverse socio-cultural contexts, covering domains such as family dynamics, civic engagement, and inequality. We further introduce the \textbf{Entire-Pipeline Survey Datasets} for task-specific supervision. At the core is \textbf{AlignSurvey-Expert (ASE)}, an expert-annotated dataset consisting of 161 semi-structured interviews and 1,679 questionnaires, comprising 2,500+ dialogues and 16,000+ responses, and enriched with detailed demographic metadata and annotated reasoning chains. To enable cross-national validation, we include two nationally representative datasets: the U.S.-based GSS and the China-based CHIP.

\begin{table*}[t]
    \centering
    \small
    \begin{tabular}{l|ccccccc}
        \toprule
        \textbf{Benchmark} & \textbf{Type} & \textbf{Size} & \textbf{Source} & \textbf{Demographic} & \textbf{Multi-Country} & \textbf{Availability} \\
        \midrule
        \textbf{Psychology-related} \\ 
        PhDGPT \cite{de2024phdgpt} & Quant & 756K & Synthetic & \checkmark & \XSolidBrush & Dataset \\
        Psych-101 \cite{binz2024centaur} & Quant & 10M & Public & \XSolidBrush & \checkmark & Model + Dataset \\
        Scenario \cite{cui2025large} & Quant & $/$ & Public & \XSolidBrush & \XSolidBrush & Dataset\\
        \midrule
        \textbf{Social Survey-related} \\
        OpinionQA \cite{Santurkar2023WhoseOD} & Quant & 80K & Public & \checkmark & \XSolidBrush & Dataset \\
        Anthology \cite{moon2024virtual} & Quant & 10K & Public & \checkmark & \XSolidBrush & Dataset \\
        SubPOP \cite{suh2025language} & Quant & 73K & Public & \checkmark &  \XSolidBrush & Dataset \\
        \textbf{AlignSurvey (Ours)} & Qual + Quant & 600K & Expert + Public & \checkmark & \checkmark  & Model + Dataset \\
        \bottomrule
    \end{tabular}
    \caption{Overview of alignment benchmark related to psychology and social surveys. “Quant” (quantitative) includes structured tasks with closed-form responses or ratings (e.g., survey choices or preference scores); “Qual” (qualitative) refers to expert-annotated open-ended responses, such as interviews and reasoning chains; and “Qual+Quant” covers both. “Source” indicates data origin: “Synthetic” is model-generated, “Public” is real-world open data, and “Expert” involves expert annotation.
}
    \label{tab:comparison}
\end{table*}

Based on AlignSurvey, we fine-tune three representative open-source LLMs to develop the \textbf{SurveyLM} model family. The training follows a two-stage alignment strategy: we first adapt base model on the Social Foundation Corpus to equip it with general social knowledge and discourse patterns, then fine-tune them on each task using the Entire-Pipeline Survey Datasets. These models serve as strong reference baselines for evaluating alignment across survey tasks.

Experiments demonstrate that general-purpose LLMs fail to reliably reproduce survey outcomes, particularly for underrepresented groups. In contrast, SurveyLM models yield substantial improvements, 10–20\% gains in demographic alignment and stance prediction, highlighting the need for domain-specific adaptation in social applications of LLMs.

Our contributions can be summarized as follows:

$\bullet$ We introduce AlignSurvey, the first benchmark that systematically replicates the full pipeline of professional social surveys using LLMs, including tasks for social role modeling, interview simulation, attitude modeling, and structured response prediction.

$\bullet$ We construct a multi-tier dataset architecture comprising (i) the Social Foundation Corpus (44K+ interviews, 400K+ survey records), and (ii) Entire-Pipeline Survey Datasets for full-pipeline alignment, enabling robust, cross-national evaluation and task-specific supervision.

$\bullet$ We release the SurveyLM model family, trained via a two-stage alignment strategy, as reference models for evaluating task-specific alignment across the survey pipeline.

$\bullet$ We contribute AlignSurvey-Expert (ASE), a multi-tiered dataset comprising expert-annotated interviews and thematic survey responses. It includes attitudinal stances, reasoning chains and demographic profiles, supporting supervision and evaluation across entire social survey process.

$\bullet$ All datasets, models, and evaluation code will be released at github and huggingface to support transparent, reproducible, and socially impactful research.
\section{Related Work}
\subsubsection{Simulating Survey Respondents with LLMs.}
To reduce the cost and rigidity of traditional survey methods~\cite{wright2010survey,heffetz2019difficulty,kalton2009methods}, recent research explores using LLMs as virtual respondents~\cite{zhou2024exploring,zhu2025llm,wang2024human,li2024political,santurkar2023whose,abdurahman2024perils}. 
Most approaches fall into prompt engineering, that simulates demographic variation via persona‐based instructions~\cite{santurkar2023whose,Hwang2023AligningLM,Simmons2022MoralML,Kim2024FewshotPO,Sun2024PersonaDBEL,moon2024virtual} and fine-tuning on real-world corpora to model individual or group preferences~\cite{Chu2023LanguageMT,He2024CommunityCrossInstructUI,Kwon2023EfficientMM,Zhao2023GroupPO,suh2025language}.
However, most of these efforts focus on isolated tasks and rarely address qualitative interviewing or full survey pipelines.

\subsubsection{Benchmarks for Aligning LLMs with Social Survey Data.}
While aligning LLMs with human preferences remains a core challenge~\cite{Kopf2023OpenAssistantC,Aroyo2023DICESDD,Lambert2024RewardBenchER,Ethayarajh2024KTOMA}, benchmarks focused on social surveys are limited. Table~\ref{tab:comparison} summarizes representative datasets across psychology and survey domains. Psychology-related resources have grown rapidly, often relying on synthetic scenarios to model cognition and responses~\cite{Binz2023TurningLL,binz2024centaur,Abdurahman2024PerilsAO,dominguez2024questioning}. However, they mostly focus on individual cognition, lack real-world demographic variation, and do not model full survey workflows. Survey-focused benchmarks such as OpinionQA, Anthology, and SubPOP~\cite{Kirk2024ThePA,Santurkar2023WhoseOD,moon2024virtual,suh2025language} target structured, fixed-option questions, typically in U.S. contexts. They overlook qualitative interviewing, reasoning chains, and cross-cultural validity. None of these benchmarks replicates the entire process of professional social surveys or enables integrated evaluation across qualitative and quantitative components with expert annotations.
\section{AlignSurvey}

\subsection{Design Principle}
AlignSurvey is designed to systematically evaluate whether large language models can replicate the entire social survey process. As shown in Figure~\ref{fig:overview}, the typical process of a professional social survey comprises four stages \cite{ahmed2024mixed,fetters2013achieving}: 
(1) Role Exploration: finding investigation targets, 
(2) Qualitative Survey: conducting semi-structured interviews, 
(3) Attitude Mining: extracting attitude stances and reasoning chains from dialogues, and 
(4) Quantitative Survey: collecting structured responses.
AlignSurvey mirrors this process into four modeling and evaluation tasks: \textit{Social Role Modeling}, \textit{Semi-structured Interview Modeling}, \textit{Attitude Stance Modeling}, and \textit{Structured Response Modeling}.
The Attitude and Response Modeling tasks include individual- and group-level subtasks to enable analysis of model alignment across multiple levels of social understanding. AlignSurvey provides stage-specific metrics to diagnose model strengths and limitations to support both rigorous benchmarking and responsible use in empirical social research.

\subsection{Data Collection and Processing}
AlignSurvey builds a multi-tiered dataset that combines large-scale public data and expert-curated resources to support comprehensive social contextual grounding and full pipeline alignment for LLM-based social surveys process.

\subsubsection{Social Foundation Corpus.} This corpus provides foundational training to equip models with broad social knowledge and cultural patterns before task-specific alignment. It comprises two components:

\textbf{\textit{Qualitative Corpus}} contains 44,021 structured interview dialogues from publicly accessible video platforms and books, spanning diverse national and cultural contexts. Dialogues are segmented into conversational turns, with a 5-turn sliding window used to predict the next response.

\textbf{\textit{Quantitative Corpus}} comprises 411,174 records from four authoritative cross-national surveys: the American Trends Panel (ATP), European Social Survey (ESS), Chinese Social Survey (CSS), and Chinese General Social Survey (CGSS). These datasets cover topics such as public opinion, trust, inequality, and civic engagement. We selected five waves of ATP (40, 41, 54, 81, 103), Round 11 of ESS, and the latest releases of CSS and CGSS. To mitigate response biases (e.g., label-position bias~\cite{dominguez2024questioning}), option labels and choices were randomly shuffled.

\subsubsection{Entire-Pipeline Survey Datasets.} As existing datasets seldom support full-pipeline alignment, we compile this suit of datasets, which comprises (1) AlignSurvey-Expert, an expert-annotated dataset, and (2) two national surveys from the U.S. and China.

\textbf{\textit{AlignSurvey-Expert (ASE)}} is the core dataset for supervising and evaluating model performance across all four survey stages. It includes:

\textbf{\textit{Qualitative Component}} includes 161 semi-structured interviews that comprise 2,500+ dialogues conducted by 15 social scientists, and each paired with rich demographic metadata (e.g., age, household size, occupation). The interviews are organized around eight core questions reflecting social perception. The first four gather general topic awareness, whereas the latter four target themes of service quality, social mobility, future expectations, and policy preferences. 
Each interview is divided into theme-specific dialogue segments, each comprising a sequence of utterances reflecting the respondent’s views. 
Each segment is annotated by six domain experts with an attitude label (positive, neutral, or negative) and a reasoning chain explaining the stance. See Appendix F for annotation details and reliability.
These chains capture underlying logic by incorporating contextual factors such as life experience, group identity, and perceived fairness. Grounded in demographic profiles and dialogue content, they enable transparent, interpretable alignment evaluation in assessing subgroup-level fidelity, and serve as key supervision signals for training attitude models. 

\textbf{\textit{Quantitative Component}} includes 1,679 questionnaires that comprise 16,000+ responses designed around the same themes as the interviews. These were collected via an online platform, with rigorous screening applied to ensure data quality. Each question contains multiple labeled answer choices (e.g., A/B/C), which were randomly shuffled to mitigate position bias. The correct answer for each question is explicitly marked to support supervised learning and evaluation in structured response modeling.

\textbf{\textit{Supplementary National Datasets.}}  
To support robustness checks and enhance cross-cultural generalization, we include two complementary national questionnaire datasets: the General Social Survey (GSS) from the U.S. and the China Household Income Project (CHIP). Both underwent the same preprocessing procedures as the questionnaire corpus of \textbf{ASE}, including label shuffling and standardized formatting, enabling consistent cross-dataset evaluation.

See Appendix A for detailed dataset information and Appendix G.1 for licensing details.

\subsection{Task Definition}
AlignSurvey defines four core tasks aligned with key stages of the social survey pipeline, each guided by tailored prompts and evaluation metrics to ensure faithful modeling and assessment.

\begin{table}[htp]
  \centering
  \small
  \begin{tabular}{llllllll}
    \toprule
    \textbf{ID} & \textbf{Task} & \textbf{Train} & \textbf{Test} \\
    \midrule
    Task1 & Social Role & 8\,712 & 2\,880 \\
    \midrule
    Task2 & Semi\textendash structured Interview  & 1\,904 & 632 \\
    \midrule
    Task3.1 & Attitude Stance(Individual) & 13\,239 & 3\,338 \\
    Task3.2 & Attitude Stance(Group) & 108 & 36 \\
    \midrule
    Task4.1 & Structured Response(Individual) & 54\,180 & 12\,047 \\
    Task4.2 & Structured Response(Group) & 46\,815 & 8\,674 \\
    \bottomrule
  \end{tabular}
  \caption{Details of tasks within AlignSurvey.}
  \label{tab:task}
\end{table}

\subsubsection{Task 1: Social Role Modeling.} This task evaluates whether an LLM can predict an interviewee’s demographic attribute (e.g., gender, education level) $c_i \in \boldsymbol{c}$ from a theme-specific dialogue segment $D^{(t)}$, where $t$ denotes the theme. 
The model prediction is:
$\hat{c_i} = f_{LLM}(D^{(t)})$. 
Model performance is measured by accuracy: the proportion of predicted $\hat{c}_i$ matching the ground-truth $c_i$:
$$
Accuracy = \frac{1}{N}\sum_{i=1}^{N}\mathbf{I}[\hat{c}_i = c_i],
$$
where \(\mathbf{I}[\cdot]\) is the indicator function.

\subsubsection{Task 2: Semi-structured Interview Modeling.} This task evaluates whether the model can generate coherent interview responses conditioned on a full demographic profile $\boldsymbol{c} = (c_1,c_2,..,c_i)$ and recent dialogue history $\mathcal{H}^{(t)}_{\tau}$, the last $\tau = 5$ utterances of a dialogue $D^{(t)}$. The model prediction is:
$\hat{d}^{(t)}_{\text{next}} = f_{LLM}(\boldsymbol{c},\mathcal{H}^{(t)}_{\tau}).$
The generated response is evaluated along three dimensions: 
(i) \textit{Naturalness} (fluency and topical coherence), 
(ii) \textit{Style Match} (alignment with interview tone), and 
(iii) \textit{Consistency} (logical and factual coherence). 

Formally, let $\mathcal{J}$ be the set of evaluators
\footnote{For Task 2 evaluation, to reduce bias and improve robustness, we use multiple strong LLMs (e.g., GPT-4o, DeepSeek-R1, Qwen-Max) as automated evaluators, with human validation confirming high agreement and consistency (see Appendix E.3).}, 
and $s^{(d)}_{j,k}$ be the score assigned by evaluator $j \in \mathcal{J}$ on dimension $k$ for dialogue $d$. The final score is computed as:
$$S^{(d)}_k = \frac{1}{|\mathcal{J}|} \sum_{j \in \mathcal{J}} s^{(d)}_{j,k}.$$

\begin{table*}[t]
  \centering
  \small
  \begin{tabular}{l|ccc|ccc|ccc}
    \toprule
    \multicolumn{1}{c|}{\textbf{Model}} &
    \multicolumn{3}{c|}{\textbf{Task 1}} &
    \multicolumn{3}{c|}{\textbf{Task 2}} &
    \multicolumn{3}{c}{\textbf{Task 3}} \\
    & Acc$_\text{Basic}\uparrow$ & Acc$_\text{Social}\uparrow$ & Acc$_\text{Family}\uparrow$ & Nat.$\uparrow$ & Sty.$\uparrow$ & Cons.$\uparrow$ & Acc.$\uparrow$ & Jaccard$\uparrow$ & WD$\downarrow$ \\
    \midrule
    GPT-4o \textit{(Zero)} & 44.08\% & 35.28\% & 3.85\% & 3.45 & 2.85 & 2.89 & 38.12\% & 0.099 & 1.648 \\
    GPT-4o \textit{(Few)} & 40.59\% & 36.58\% & 17.43\% & 3.49 & 2.80 & 2.75 & 42.73\% & 0.103 & 1.180 \\
    Claude 3.7 Sonnet \textit{(Zero)} & 42.72\% & 34.17\% & 4.06\% & 3.41 & 2.81 & 2.85 & 28.12\% & 0.064 & 1.332 \\
    Claude 3.7 Sonnet \textit{(Few)} & 40.15\% & 33.18\% & 14.34\% & 3.48 & 2.83 & 2.90 & 31.48\% & 0.072 & 1.120 \\
    DeepSeek‑R1 \textit{(Zero)} & 38.52\% & 34.31\% & 6.56\% & 3.40 & 2.81 & 2.83 & 36.88\% & 0.075 & 1.356 \\
    DeepSeek‑R1 \textit{(Few)} & 36.36\% & 30.53\% & 18.27\% & 3.47 & 2.70 & 2.78 & 41.05\% & 0.081 & 1.100 \\
    R1‑Distill‑Qwen‑14B \textit{(Zero)} & 30.25\% & 27.08\% & 2.81\% & 2.78 & 2.20 & 2.36 & 38.75\% & 0.105 & 1.795 \\
    R1‑Distill‑Qwen‑14B \textit{(Few)} & 34.72\% & 31.18\% & 15.87\% & 2.85 & 2.25 & 2.30 & 33.81\% & 0.111 & 1.140 \\
    Qwen2.5‑72B \textit{(Zero)} & 37.83\% & 33.34\% & 3.12\% & 3.40 & 2.72 & 2.85 & 35.00\% & 0.102 & 1.653 \\
    Qwen2.5‑72B \textit{(Few)} & 38.65\% & 32.75\% & 16.90\% & 3.46 & 2.65 & 2.70 & 42.16\% & 0.113 & 1.050 \\
    \midrule

    Mistral‑7B‑v0.3 \textit{(Zero)} & 32.58\% & 35.14\% & 5.10\% & 3.21 & 2.57 & 2.44 & 38.44\% & 0.106 & 1.534 \\
    Mistral‑7B‑v0.3 \textit{(Few)} & 39.31\% & 32.73\% & 16.53\% & 3.28 & 2.50 & 2.40 & 39.77\% & 0.114 & 0.302 \\
    Meta‑Llama‑3.1‑8B \textit{(Zero)} & 37.00\% & 33.89\% & 8.33\% & 3.30 & 2.69 & 2.79 & 39.38\% & 0.107 & 1.521 \\
    Meta‑Llama‑3.1‑8B \textit{(Few)} & 35.42\% & 32.52\% & 16.02\% & 3.36 & 2.60 & 2.68 & 39.96\% & 0.119 & 0.447 \\
    Qwen2.5‑7B \textit{(Zero)} & 36.08\% & 33.89\% & 2.71\% & 3.39 & 2.67 & 2.79 & 36.88\% & 0.104 & 1.700 \\
    Qwen2.5‑7B \textit{(Few)} & 27.92\% & 37.83\% & 16.76\% & 3.44 & 2.60 & 2.85 & 44.69\% & 0.118 & 0.372 \\
    \midrule
     
    $SurveyLM_{\text{Mistral‑7B‑v0.3}}$ & 48.24\%$^{**}$ & 52.23\%$^{**}$ & 47.08\%$^{**}$ & 2.61 & 2.90$^{*}$ & 2.90$^{*}$ & \textbf{57.13\%$^{**}$} & 0.131$^{**}$ & \textbf{0.297$^{**}$} \\
    $SurveyLM_\text{Meta‑Llama‑3.1‑8B}$ & 50.55\%$^{**}$ & \textbf{58.32\%$^{**}$} & 44.73\%$^{**}$ & 3.77$^{*}$ & \textbf{3.00$^{*}$} & 2.94$^{*}$ & 55.09\%$^{**}$ & \textbf{0.134$^{**}$} & 0.458$^{**}$ \\
    $SurveyLM_\text{Qwen2.5‑7B}$ & \textbf{54.65\%$^{**}$} & 57.23\%$^{**}$ & \textbf{48.71\%$^{**}$} & \textbf{3.98$^{*}$} & 2.96$^{*}$ & \textbf{3.01$^{*}$} & 56.83\%$^{**}$ & 0.128$^{**}$ & 0.385$^{**}$ \\
    \bottomrule
  \end{tabular}
  \caption{Multi‑Task Evaluation Results. Even strong baselines such as GPT-4o still struggle, yielding low performance on fine-grained demographic prediction, interview style recognition, and stance detection. SurveyLM raises accuracy by about 10 to 15 percentage points and roughly halves Wasserstein distance. We conducted t-tests between each SurveyLM and its zero-shot model; $^{*}p<0.05$ and $^{**}p<0.01$. $\uparrow$ indicate that higher values are better, while $\downarrow$ indicate that lower values are better.}
  \label{tab:task123}
\end{table*}

\subsubsection{Task 3: Attitude Stance Modeling.} This task involves predicting both individual- and group-level attitude stances $\mathcal{A}$ and reasoning chains $\mathcal{R}$.

\textbf{\textit{Individual-level.}}
Given individual demographic profile $\boldsymbol{c}$ and theme $t$, the model predicts the attitude stance and reasoning chain:
$(\hat{a}^{(t)}, \hat{r}^{(t)}) = f_{LLM}(\boldsymbol{c}, t).$
Predictions are evaluated using macro-averaged accuracy, precision, recall, and F1 (for attitude stances), and ROUGE, Jaccard, and cosine similarity (for reasoning chains).

\textbf{\textit{Group-level.}}
For each question, the model produces individual stance predictions ${\hat{a}i^{(t)}}{i=1}^N$ for respondents belonging to a demographic group $t$. We convert these predictions into an empirical stance distribution by counting label frequencies: $
\hat{\mathbf{p}} = \text{Aggregate}(\{\hat{a}_i^{(t)}\}), \quad \hat{\mathbf{p}} \in \Delta^{|\mathcal{A}| - 1}.
$ The predicted distribution is compared with the reference distribution $\mathbf{p}$ using the Wasserstein distance:
$$
W_1(\hat{\mathbf{p}}, \mathbf{p}) = \min_{\Gamma \ge 0, \Gamma \mathbf{1} = \hat{\mathbf{p}}, \Gamma^\top \mathbf{1} = \mathbf{p}} \sum_{u,v \in \mathcal{A}} \Gamma_{uv} C_{uv},
$$
where $C_{uv}$ denotes the ground distance between stance labels $u, v \in \{\text{positive}, \text{neutral}, \text{negative}\}$.

To mitigate bias, we ground predictions in expert-annotated reasoning chains and training with demographic fine-grained profiles. This enables interpretable, group-aware alignment without relying on stereotypes.

\subsubsection{Task 4: Survey Response Modeling.}This task predicts individual and group-level responses to questionnaire items.

\textbf{\textit{Individual-level.}}
Given a demographic profile $\boldsymbol{c}$ and question $Q_i$ with labeled options $(l_{i}, o_{i})$, the model predicts the respondent's selected answer $
(\hat{l}_i, \hat{o}_i) = f_{LLM}(\boldsymbol{c}, Q_i)
$, where $l_i$ denotes the label (e.g., A/B…) and $o_i$ denotes the content.
Predictions are evaluated using accuracy, macro-averaged precision, recall, and F1, based on the ground-truth answers.

\textbf{\textit{Group-level.}}
Aggregating individual responses, the model generates a distribution over possible answers:
$
\hat{\mathbf{p}}_q = \text{Aggregate}(\{\hat{o}_{i}\}), \quad \hat{\mathbf{p}}_q \in \Delta^{|L|-1}.
$
Alignment is measured via the Wasserstein distance (Similar to Task 3).

Table~\ref{tab:task} summarizes the task definitions and dataset statistics. Further task-specific details and examples are provided in the Appendix D and E.

\subsubsection{Constructing the SurveyLM Family.}
We adopt a two-stage alignment strategy to build the \textbf{SurveyLM} model family. First, we fine-tune three open-source LLMs (Mistral 7B, LLaMA 3.1 8B, Qwen 2.5 7B) on the Social Foundation Corpus to equip them with general social concepts and culturally diverse discourse patterns. Then, we fine-tune the adapted models on the four AlignSurvey tasks using the Entire-Pipeline Survey Datasets, enabling task-specific alignment with both qualitative and quantitative objectives. The resulting models serve as strong reference baselines for domain-aligned social survey modeling and will be publicly released to support reproducible research.
\begin{table*}[htp]
  \centering
  \small
  \newcolumntype{C}{>{\centering\arraybackslash}p{1cm}}
  \begin{tabular}{l|CCC|CCC|CCC}
    \toprule
    \multicolumn{1}{c|}{\textbf{Model}} &
    \multicolumn{3}{c|}{\textbf{ASE}} &
    \multicolumn{3}{c|}{\textbf{CHIP}} &
    \multicolumn{3}{c}{\textbf{GSS}} \\
    & Acc.$\uparrow$ & F1$\uparrow$ & WD$\downarrow$ & Acc.$\uparrow$ & F1$\uparrow$ & WD$\downarrow$ & Acc.$\uparrow$ & F1$\uparrow$ & WD$\downarrow$ \\
    \midrule
    GPT-4o \textit{(Zero)} & 46.56\% & 22.05\% & 2.1336 & 53.43\% & 26.24\% & 1.9325 & 32.79\% & 24.41\% & 1.1130 \\
    GPT-4o \textit{(Few)}  & 47.91\% & 23.10\% & 1.7895 & 54.32\% & 27.30\% & 1.6992 & 33.85\% & 25.10\% & 1.0284 \\
    Claude 3.7 Sonnet \textit{(Zero)} & 42.11\% & 18.44\% & 1.7789 & 47.21\% &  9.50\% & 1.8872 & 33.94\% & 25.93\% & 1.2248 \\
    Claude 3.7 Sonnet \textit{(Few)}  & 42.95\% & 19.50\% & 1.6542 & 47.98\% &  9.95\% & 1.7318 & 34.55\% & 26.60\% & 1.1095 \\
    DeepSeek‑R1 \textit{(Zero)} & 47.21\% &  7.11\% & 1.9987 & 44.65\% &  4.52\% & 1.7789 & 34.54\% & 23.15\% & 1.3854 \\
    DeepSeek‑R1 \textit{(Few)}  & 47.88\% &  7.65\% & 1.8527 & 45.27\% &  4.90\% & 1.6523 & 35.20\% & 23.70\% & 1.2432 \\
    R1‑Distill‑Qwen‑14B \textit{(Zero)} & 49.54\% & 25.90\% & 1.9295 & 50.62\% & 29.36\% & 1.9448 & 29.39\% & 11.57\% & 0.7920 \\
    R1‑Distill‑Qwen‑14B \textit{(Few)}  & 50.29\% & 26.80\% & 1.7164 & 51.37\% & 30.67\% & 1.7104 & 30.20\% & 12.25\% & 0.7548 \\
    Qwen2.5‑72B \textit{(Zero)} & 42.20\% &  3.80\% & 1.8654 & 47.98\% & 13.29\% & 1.9918 & 34.50\% & 21.14\% & 1.2180 \\
    Qwen2.5‑72B \textit{(Few)}  & 43.06\% &  4.05\% & 1.7348 & 48.66\% & 14.02\% & 1.7989 & 35.60\% & 22.05\% & 1.1112 \\
    \midrule
    Mistral‑7B‑v0.3 \textit{(Zero)} & 28.04\% &  1.10\% & 2.2118 & 23.60\% &  3.46\% & 1.9941 &  3.19\% &  0.20\% & 1.1479 \\
    Mistral‑7B‑v0.3 \textit{(Few)}  & 30.47\% &  1.25\% & 2.0823 & 26.89\% &  3.70\% & 1.8876 & 10.30\% &  0.24\% & 1.0913 \\
    Meta‑Llama‑3.1‑8B \textit{(Zero)} & 14.84\% &  0.17\% & 2.1154 & 20.84\% &  2.40\% & 1.7748 &  1.61\% &  0.08\% & 1.5004 \\
    Meta‑Llama‑3.1‑8B \textit{(Few)}  & 17.26\% &  0.19\% & 1.9451 & 22.46\% &  2.60\% & 1.6435 &  6.90\% &  0.12\% & 1.3817 \\
    Qwen2.5‑7B \textit{(Zero)} & 16.39\% &  0.12\% & 1.8479 & 51.95\% & 12.51\% & 1.9099 & 25.33\% &  4.30\% & 1.2411 \\
    Qwen2.5‑7B \textit{(Few)}  & 18.46\% &  0.15\% & 1.7059 & 53.14\% & 13.15\% & 1.7441 & 26.85\% &  4.55\% & 1.1371 \\
    \midrule
    $SurveyLM_{\text{Mistral‑7B‑v0.3}}$ & 54.59\%$^{**}$ & 41.34\%$^{**}$ & 0.0671$^{**}$ & 65.56\%$^{**}$ & \textbf{35.35\%$^{**}$} & 0.0857$^{**}$ & 51.06\%$^{**}$ & 35.73\%$^{**}$ & 0.2097$^{**}$ \\
    $SurveyLM_{\text{Meta‑Llama‑3.1‑8B}}$ & \textbf{56.11\%$^{**}$} & 41.47\%$^{**}$ & 0.0675$^{**}$ & 64.89\%$^{**}$ & 32.99\%$^{**}$ & 0.0845$^{**}$ & \textbf{51.44\%$^{**}$} & 33.84\%$^{**}$ & 0.2137$^{**}$ \\
    $SurveyLM_{\text{Qwen2.5‑7B}}$ & 55.47\%$^{**}$ & \textbf{41.79\%$^{**}$} & \textbf{0.0541$^{**}$} & \textbf{67.17\%$^{**}$} & 34.94\%$^{**}$ & \textbf{0.0686$^{**}$} & 51.52\%$^{**}$ & \textbf{36.22\%$^{**}$} & \textbf{0.1849$^{**}$} \\
    \bottomrule
  \end{tabular}
  \caption{Survey Response Modeling (Task4) Results. In zero- and few-shot settings, general-purpose LLMs give low-fidelity answers. On the ASE, CHIP, and GSS datasets, SurveyLM achieves high accuracy and F1 scores while yielding substantially smaller Wasserstein distances. The large F1 jump shows SurveyLM no longer defaults to midpoint guesses but distinguishes the full spectrum, reducing central-tendency bias. Two-sided t-tests vs. zero-shot baselines: $^{*}p<0.05$, $^{**}p<0.01$. $\uparrow$ indicate that higher values are better, while $\downarrow$ indicate that lower values are better.}
  \label{tab:task4}
\end{table*}

\section{Experiments \& Discussion}

\subsection{Setup and Implementation Details}

We evaluate models under three settings: \textbf{zero-shot}, \textbf{few-shot} (with three in-context examples), and \textbf{supervised fine-tuning (SFT)}. SFT corresponds to the training procedure used to construct the \textbf{SurveyLM} family, which adapts LLMs through corpus-level and task-specific supervision. SFT first trains on the Social Foundation Corpus (1 epoch), followed by each task dataset (3 epochs). Prompts exceeding the model’s context window are truncated by removing the middle segment, retaining the task instruction and query. Prompt templates are provided in Appendix D and E.

We apply LoRA (rank 8) targeting all attention and feed-forward layers. Training uses AdamW with learning rate $1\times10^{-4}$, cosine decay schedule, 10\% linear warm-up, and bfloat16 precision. The effective batch size is 32 (per-device batch size of 4 with gradient accumulation of 8). All experiments fix the random seed to 42. Inference runs on 8$\times$ NVIDIA H20 GPUs or official APIs for proprietary models.

\subsection{Evaluated Models}
We benchmark three groups of models for comparative analysis.
\textbf{Representative Top-tier Models:} GPT-4o, Claude 3.7 Sonnet, DeepSeek R1, DeepSeek R1 Distill Qwen 14B, and Qwen 2.5 72B,  serve as references for general-purpose performance.
\textbf{Open-source Base Models:} Meta LLaMA 3.1 8B, Qwen 2.5 7B, and Mistral 7B v0.3 act as task-agnostic baselines, allowing analysis of architecture and scale effects on alignment.
\textbf{SurveyLM Family:} SurveyLM$_{\text{Meta-LLaMA-3.1-8B}}$, SurveyLM$_{\text{Qwen 2.5-7B}}$, and SurveyLM$_{\text{Mistral-7B-v0.3}}$ specialized for the full survey pipeline and used to evaluate supervised domain alignment.

\subsection{Experimental Results \& Analysis}

\begin{figure}[htp]
    \centering
    \includegraphics[width=\linewidth]{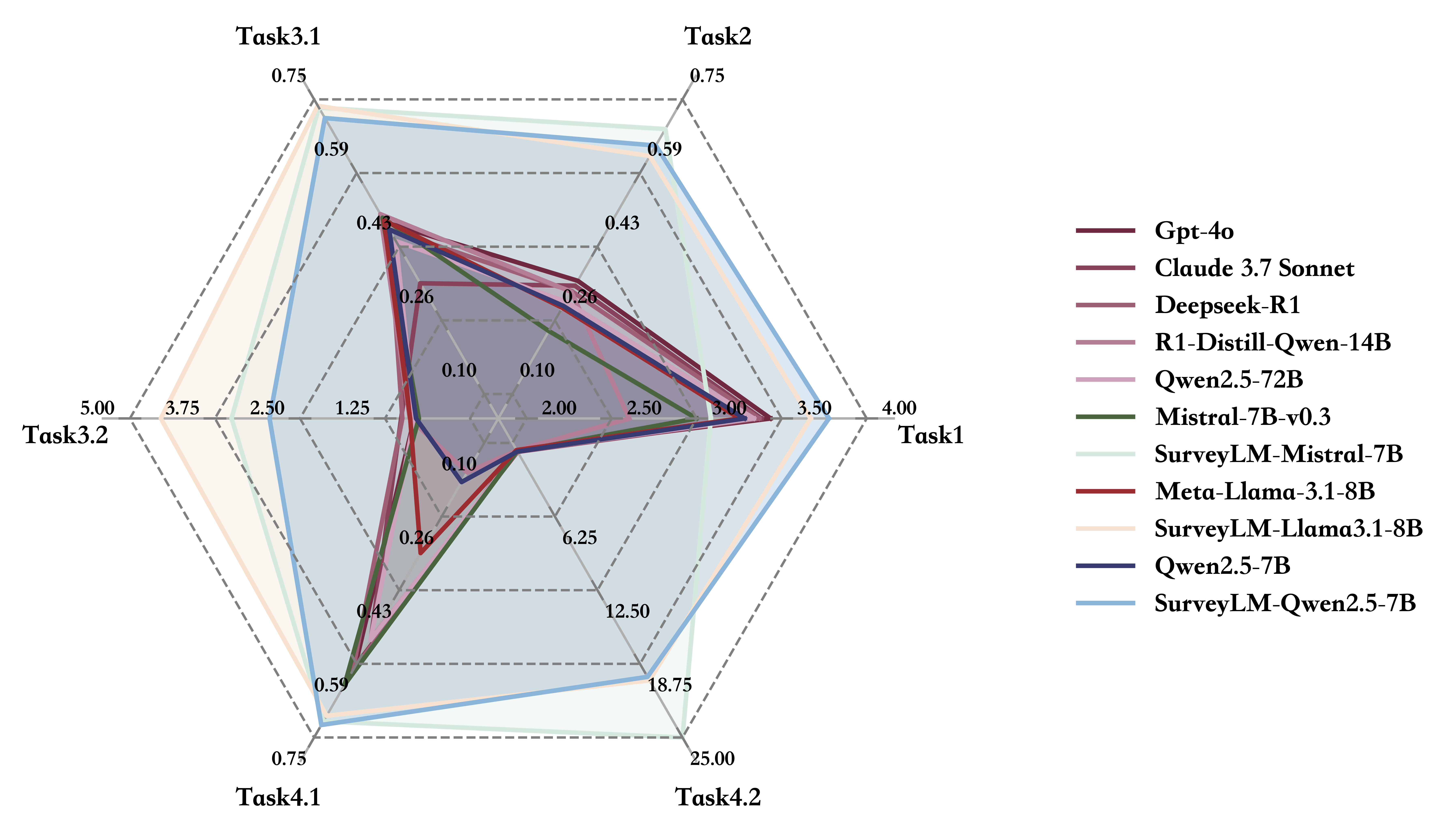}
    \caption{A radar chart showing the overall performance on the tasks. Each axis represents the task score, with larger values indicating better performance. We used the inverse of the WD. Each broken line corresponds to a model, with larger enclosed areas indicating better overall performance. Overall, SurveyLM series performed exceptionally well.}
    \label{fig:rander}
\end{figure}

Table~\ref{tab:task123} and Table~\ref{tab:task4} report aggregated results for Tasks 1–3 and Task 4. For Task 1, we divided accuracy into three categories which specific classification can be found in Appendix B.1. Figure~\ref{fig:rander} complements these with a radar chart overview of model performance across tasks. Detailed per-label scores and metrics are provided in Appendix B and C.

\subsubsection{Task 1: Social Role Modeling.}
SurveyLM models significantly outperform all baselines in predicting fine-grained demographic attributes. While top-tier models like GPT-4o perform decently on binary traits (e.g., gender), they struggle with complex categories such as family structure and social roles. Few-shot prompting yields limited gains. SurveyLM models improve accuracy by over 10 points across attributes, with the strongest gains in familial and social context fields. Notably, a fine-tuned 7B model (Qwen) surpasses its 72B counterpart, highlighting the impact of domain alignment over model scale.

\subsubsection{Task 2: Semi-structure Interview Modeling.}
General-purpose models (e.g., GPT-4o, DeepSeek-R1) reach moderate naturalness but lack stylistic consistency and fail to match the interview style. SurveyLM models, with two-stage alignment, yield more fluent, coherent, and contextually appropriate responses, demonstrating the importance of fine-tuning for qualitative generation.

\subsubsection{Task3. Attitude Stance Modeling.} 
General-purpose models tend to default to majority stances with generic reasoning, showing 28–45\% accuracy and limited few-shot improvement. They perform poorly on group-level metrics Wasserstein Distance. SurveyLM models exceed 55\% accuracy and generate higher-quality explanations (Jaccard higher), confirming the value of alignment for accurate, interpretable attitude inference in public opinion modeling.

\subsubsection{Task4. Survey Response Modeling.} In both zero- and few-shot settings, general-purpose models yield low-fidelity outputs (accuracy $<$50\%, low F1). SurveyLM achieves high accuracy and F1 scores while yielding substantially
smaller Wasserstein distances on ASE, CHIP, and GSS. The pronounced jump in F1 shows that the model no longer “plays it safe” by predicting the midpoint response for everyone. It learns to recognize and output the full spectrum of options. This balanced improvement mitigates the \textbf{central-tendency bias} that questionnaires often suffer from. These results demonstrate the strength of supervised alignment for structured response prediction.
\subsubsection{Ablation Analysis.}
To assess the impact of each alignment stage, we conduct ablation tests on three backbones (Mistral-7B, LLaMA-3.1-8B, Qwen2.5-7B), removing either Stage I (foundation adaptation, \textit{w/o Foundation}) or Stage II (task-specific tuning, \textit{w/o Task SFT}), keeping other conditions fixed.

\begin{table}[htp]
  \centering
  \small
  \newcolumntype{C}{>{\centering\arraybackslash}p{0.5cm}}
  \newcolumntype{P}{>{\centering\arraybackslash}p{0.2cm}}
  \begin{tabular}{lPPCCCC}
    \toprule
    \multicolumn{1}{c}{\textbf{Model}} &
    \multicolumn{1}{c}{\textbf{T1}} &
    \multicolumn{1}{c}{\textbf{T2}} &
    \multicolumn{2}{c}{\textbf{T3}} &
    \multicolumn{2}{c}{\textbf{T4(ASE)}} \\
    \cmidrule(r){4-5}
    \cmidrule(r){6-7}
    & Acc & Avg & Acc & WD & Acc & WD \\
    \midrule
    $SurveyLM_\text{Mistral}$ & \textbf{0.47} & 2.80 & \textbf{0.57} & \textbf{0.297} & \textbf{0.55} & \textbf{0.067}\\
    \textit{ \ \ w/o Foundation} & 0.46 & \textbf{3.10}& 0.56 & 0.303 & 0.53 & 0.077\\
    \textit{ \ \ w/o Task SFT} & 0.38 & 2.96 & 0.44 & 0.313 & 0.36 & 2.182 \\
    $SurveyLM_\text{Llama}$ & \textbf{0.45} & \textbf{3.24} & \textbf{0.55} & \textbf{0.458} & \textbf{0.56} & \textbf{0.068} \\
    \textit{ \ \ w/o Foundation} & 0.44 & 3.07 & 0.54 & 0.463 & 0.55 & 0.085 \\
    \textit{ \ \ w/o Task SFT} & 0.39 & 3.01 & 0.36 & 0.484 & 0.34 & 2.098 \\
    $SurveyLM_\text{Qwen}$ & \textbf{0.49} & \textbf{3.32} & \textbf{0.57} & \textbf{0.385} & \textbf{0.56}  & \textbf{0.054} \\
    \textit{ \ \ w/o Foundation} & 0.47 & 3.12 & 0.56 & 0.389 & 0.55 & 0.889 \\
    \textit{ \ \ w/o Task SFT} & 0.35 & 2.79 & 0.47 & 0.397 & 0.36 & 2.002 \\
    \bottomrule
  \end{tabular}
  \caption{Ablation results on Tasks 1–4. T1–T4 correspond to Tasks 1–4, respectively. For Tasks 1 and 2 we report accuracy and the average score; for Task 4 we report results on the ASE split only. Detailed results appear in Appendix C.1.}
  \label{tab:Ablation}
\end{table}

As shown in Table~\ref{tab:Ablation}, removing either stage degrades performance. Omitting foundation adaptation yields moderate drops, while skipping task-specific tuning causes severe degradation. Full SurveyLM models outperform ablated variants, confirming that foundation-level pretraining and task-specific supervision are complementary. This highlights the effectiveness of our two-stage alignment strategy in adapting general-purpose LLMs to social survey tasks.

\begin{figure}[t]
    \centering
    \includegraphics[width=\linewidth]{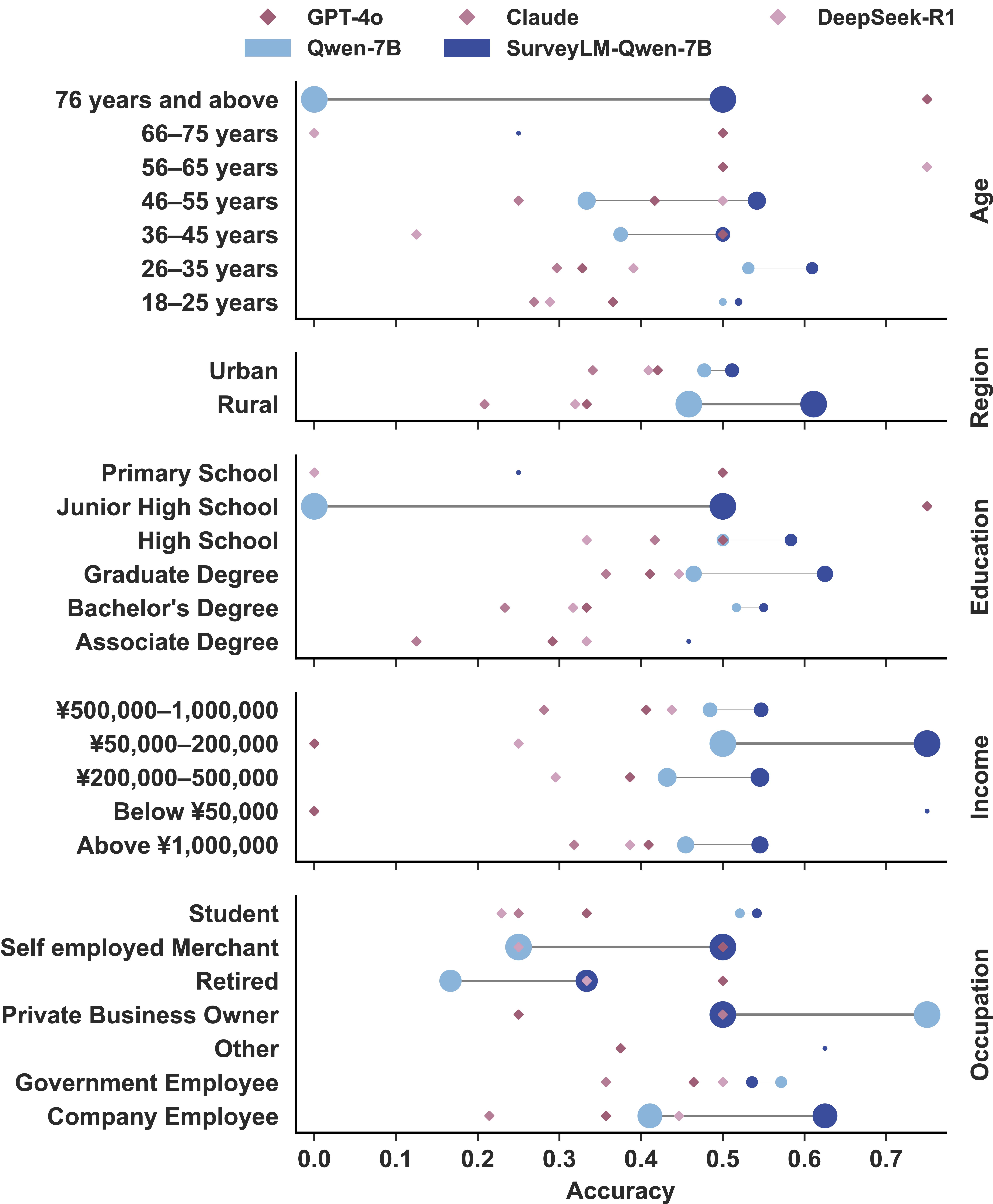}
    \caption{Multi‑Demographic Accuracy Comparison. Circle size represents the accuracy gap between Qwen-7B and its SurveyLM. Gains are especially pronounced for under‑represented groups (rural residents, aged 76+, self‑employed, low‑ and middle‑ income groups).}
    \label{fig:task3_ana}
\end{figure}

\subsubsection{Equity-Oriented Gains.}
Figure~\ref{fig:task3_ana} shows Task 3 accuracy across demographic attributes using the Qwen2.5-7B. SurveyLM consistently shifts scores toward higher accuracy, while baselines often remain near or below midpoint. Notably, SurveyLM significantly boosts accuracy for \textbf{underrepresented groups}, including rural, elderly (76+), self-employed, and low-/middle-income individuals, reducing disparities with advantaged counterparts (urban, college-educated, high-income) and enhancing demographic equity. Full comparisons across models are in the Appendix C.2. 

\subsection{Social Impact and Alignment Implications}
Our experiments show that socially grounded alignment enhances both performance and fairness. AlignSurvey and SurveyLM help recover signals from marginalized groups, reducing reliance on dominant narratives and supporting more inclusive policy diagnostics, welfare targeting, and equitable decision-making. More broadly, this work demonstrates how targeted alignment can bridge LLMs with real-world societal needs,especially in governance, auditing, and digital public services.
\section{Conclusion \& Future Work}

This paper presents AlignSurvey, the first benchmark to systematically replicate the full pipeline of social surveys using LLMs. By integrating Social Foundation Corpus and Entire-Pipeline Survey Datasets, it enables comprehensive evaluation across demographic modeling, qualitative interaction, attitude inference, and structured response prediction. Experiments demonstrate alignment on AlignSurvey can recover signals from underrepresented groups, reducing bias and supporting more inclusive, policy-relevant modeling.

Future directions include iterative improvement through human-in-the-loop feedback and expanded coverage across diverse cultural settings for broader applicability.

\section{Ethical Statement}
We take ethics seriously. Our dataset is sourced from publicly accessible content, including video-platform APIs, research-cleared books, and licensed surveys. Personally identifiable information is removed or anonymized before training, and only privacy-preserving, non-harmful content is released. Raw audio, video, or verbatim transcripts are never shared. 

We implement multiple safeguards to prevent misuse, including controlled data access, content filtering, and usage guidance tailored to responsible LLM-based survey applications. Data is used solely for non-commercial academic research, and compliance with privacy and copyright regulations is continuously monitored. See Appendix~G for further details.

\section{Acknowledgments}
This research was supported by the National Natural Science Foundation of China (Grant Nos. 72574198, 72434004), the Key Project of Humanities and Social Sciences of the Ministry of Education of China (Grant No. 2023JZDZ038), and the National Social Science Fund of China (Grant No. 23BZZ088). We thank all collaborators contributing to data collection, annotations and valuable feedback.

\bibliography{aaai2026}

@article{heffetz2019difficulty,
  title={Difficulty of reaching respondents and nonresponse Bias: Evidence from large government surveys},
  author={Heffetz, Ori and Reeves, Daniel B},
  journal={Review of Economics and Statistics},
  volume={101},
  number={1},
  pages={176--191},
  year={2019},
  publisher={MIT Press One Rogers Street, Cambridge, MA 02142-1209, USA journals-info~…}
}

@article{moy2016problems,
  title={Problems and prospects in survey research},
  author={Moy, Patricia and Murphy, Joe},
  journal={Journalism \& Mass Communication Quarterly},
  volume={93},
  number={1},
  pages={16--37},
  year={2016},
  publisher={SAGE Publications Sage CA: Los Angeles, CA}
}

@article{evans2018value,
  title={The value of online surveys: A look back and a look ahead},
  author={Evans, Joel R and Mathur, Anil},
  journal={Internet research},
  volume={28},
  number={4},
  pages={854--887},
  year={2018},
  publisher={Emerald Publishing Limited}
}

@article{prosser2018twilight,
  title={The twilight of the polls? A review of trends in polling accuracy and the causes of polling misses},
  author={Prosser, Christopher and Mellon, Jonathan},
  journal={Government and Opposition},
  volume={53},
  number={4},
  pages={757--790},
  year={2018},
  publisher={Cambridge University Press}
}

@article{wright2010survey,
  title={Survey research and social science: History, current practice, and future prospects},
  author={Wright, James D and Marsden, Peter V and others},
  journal={Handbook of survey research},
  pages={3--26},
  year={2010}
}

@article{tourangeau2004survey,
  title={Survey research and societal change},
  author={Tourangeau, Roger},
  journal={Annu. Rev. Psychol.},
  volume={55},
  number={1},
  pages={775--801},
  year={2004},
  publisher={Annual Reviews}
}

@article{kalton2009methods,
  title={Methods for oversampling rare subpopulations in social surveys},
  author={Kalton, Graham},
  journal={Survey methodology},
  volume={35},number={2},
  pages={125--141},
  year={2009}
}

@article{Hwang2023AligningLM,
  title={Aligning Language Models to User Opinions},
  author={EunJeong Hwang and Bodhisattwa Prasad Majumder and Niket Tandon},
  journal={ArXiv},
  year={2023},
  volume={abs/2305.14929},
  url={https://api.semanticscholar.org/CorpusID:258865429}
}

@article{Simmons2022MoralML,
  title={Moral Mimicry: Large Language Models Produce Moral Rationalizations Tailored to Political Identity},
  author={Gabriel Simmons},
  journal={ArXiv},
  year={2022},
  volume={abs/2209.12106},
  url={https://api.semanticscholar.org/CorpusID:252531526}
}

@article{de2024phdgpt,
  title={Phdgpt: Introducing a psychometric and linguistic dataset about how large language models perceive graduate students and professors in psychology},
  author={De Duro, Edoardo Sebastiano and Taietta, Enrique and Improta, Riccardo and Stella, Massimo},
  journal={arXiv preprint arXiv:2411.10473},
  year={2024}
}

@article{lee2024can,
  title={Can large language models estimate public opinion about global warming? An empirical assessment of algorithmic fidelity and bias},
  author={Lee, Sanguk and Peng, Tai-Quan and Goldberg, Matthew H and Rosenthal, Seth A and Kotcher, John E and Maibach, Edward W and Leiserowitz, Anthony},
  journal={PLOS Climate},
  volume={3},
  number={8},
  pages={e0000429},
  year={2024},
  publisher={Public Library of Science}
}

@article{moon2024virtual,
  title={Virtual personas for language models via an anthology of backstories},
  author={Moon, Suhong and Abdulhai, Marwa and Kang, Minwoo and Suh, Joseph and Soedarmadji, Widyadewi and Behar, Eran Kohen and Chan, David M},
  journal={arXiv preprint arXiv:2407.06576},
  year={2024}
}

@article{abdurahman2024perils,
  title={Perils and opportunities in using large language models in psychological research},
  author={Abdurahman, Suhaib and Atari, Mohammad and Karimi-Malekabadi, Farzan and Xue, Mona J and Trager, Jackson and Park, Peter S and Golazizian, Preni and Omrani, Ali and Dehghani, Morteza},
  journal={PNAS nexus},
  volume={3},
  number={7},
  pages={pgae245},
  year={2024},
  publisher={Oxford University Press US}
}

@article{suh2025language,
  title={Language model fine-tuning on scaled survey data for predicting distributions of public opinions},
  author={Suh, Joseph and Jahanparast, Erfan and Moon, Suhong and Kang, Minwoo and Chang, Serina},
  journal={arXiv preprint arXiv:2502.16761},
  year={2025}
}

@article{zhang2025socioverse,
  title={Socioverse: A world model for social simulation powered by llm agents and a pool of 10 million real-world users},
  author={Zhang, Xinnong and Lin, Jiayu and Mou, Xinyi and Yang, Shiyue and Liu, Xiawei and Sun, Libo and Lyu, Hanjia and Yang, Yihang and Qi, Weihong and Chen, Yue and others},
  journal={arXiv preprint arXiv:2504.10157},
  year={2025}
}

@inproceedings{santurkar2023whose,
  title={Whose opinions do language models reflect?},
  author={Santurkar, Shibani and Durmus, Esin and Ladhak, Faisal and Lee, Cinoo and Liang, Percy and Hashimoto, Tatsunori},
  booktitle={International Conference on Machine Learning},
  pages={29971--30004},
  year={2023},
  organization={PMLR}
}

@article{zhou2024exploring,
  title={Exploring the application of LLM-based AI in UX design: an empirical case study of ChatGPT},
  author={Zhou, Zhibin and Li, Yaoqi and Yu, Junnan},
  journal={Human--Computer Interaction},
  pages={1--33},
  year={2024},
  publisher={Taylor \& Francis}
}

@inproceedings{zhu2025llm,
  title={A llm-based controllable, scalable, human-involved user simulator framework for conversational recommender systems},
  author={Zhu, Lixi and Huang, Xiaowen and Sang, Jitao},
  booktitle={Proceedings of the ACM on Web Conference 2025},
  pages={4653--4661},
  year={2025}
}

@inproceedings{wang2024human,
  title={Human-llm collaborative annotation through effective verification of llm labels},
  author={Wang, Xinru and Kim, Hannah and Rahman, Sajjadur and Mitra, Kushan and Miao, Zhengjie},
  booktitle={Proceedings of the 2024 CHI Conference on Human Factors in Computing Systems},
  pages={1--21},
  year={2024}
}

@article{li2024political,
  title={Political-llm: Large language models in political science},
  author={Li, Lincan and Li, Jiaqi and Chen, Catherine and Gui, Fred and Yang, Hongjia and Yu, Chenxiao and Wang, Zhengguang and Cai, Jianing and Zhou, Junlong Aaron and Shen, Bolin and others},
  journal={arXiv preprint arXiv:2412.06864},
  year={2024}
}

@article{thapa2025large,
  title={Large language models (llm) in computational social science: prospects, current state, and challenges},
  author={Thapa, Surendrabikram and Shiwakoti, Shuvam and Shah, Siddhant Bikram and Adhikari, Surabhi and Veeramani, Hariram and Nasim, Mehwish and Naseem, Usman},
  journal={Social Network Analysis and Mining},
  volume={15},
  number={1},
  pages={1--30},
  year={2025},
  publisher={Springer}
}

@article{mellon2024ais,
  title={Do AIs know what the most important issue is? Using language models to code open-text social survey responses at scale},
  author={Mellon, Jonathan and Bailey, Jack and Scott, Ralph and Breckwoldt, James and Miori, Marta and Schmedeman, Phillip},
  journal={Research \& Politics},
  volume={11},
  number={1},
  pages={20531680241231468},
  year={2024},
  publisher={SAGE Publications Sage UK: London, England}
}

@inproceedings{giorgi2025human,
  title={Human and LLM biases in hate speech annotations: A socio-demographic analysis of annotators and targets},
  author={Giorgi, Tommaso and Cima, Lorenzo and Fagni, Tiziano and Avvenuti, Marco and Cresci, Stefano},
  booktitle={Proceedings of the International AAAI Conference on Web and Social Media},
  volume={19},
  pages={653--670},
  year={2025}
}

@article{abeliuk2025fairness,
  title={Fairness in LLM-Generated Surveys},
  author={Abeliuk, Andr{\'e}s and Gaete, Vanessa and Bro, Naim},
  journal={arXiv preprint arXiv:2501.15351},
  year={2025}
}

@article{wang2025large,
  title={Large language models that replace human participants can harmfully misportray and flatten identity groups},
  author={Wang, Angelina and Morgenstern, Jamie and Dickerson, John P},
  journal={Nature Machine Intelligence},
  pages={1--12},
  year={2025},
  publisher={Nature Publishing Group UK London}
}

@article{hu2025generative,
  title={Generative language models exhibit social identity biases},
  author={Hu, Tiancheng and Kyrychenko, Yara and Rathje, Steve and Collier, Nigel and van der Linden, Sander and Roozenbeek, Jon},
  journal={Nature Computational Science},
  volume={5},
  number={1},
  pages={65--75},
  year={2025},
  publisher={Nature Publishing Group US New York}
}

@article{ahmed2024mixed,
  title={Mixed methods research: Combining both qualitative and quantitative approaches},
  author={Ahmed, Amina and Pereira, Lucas and Jane, K},
  journal={en. In: ResearchGate (Sept. 2024)},
  pages={1--10},
  year={2024}
}

@article{fetters2013achieving,
  title={Achieving integration in mixed methods designs—principles and practices},
  author={Fetters, Michael D and Curry, Leslie A and Creswell, John W},
  journal={Health services research},
  volume={48},
  number={6pt2},
  pages={2134--2156},
  year={2013},
  publisher={Wiley Online Library}
}

@article{hofmann2024ai,
  title={AI generates covertly racist decisions about people based on their dialect},
  author={Hofmann, Valentin and Kalluri, Pratyusha Ria and Jurafsky, Dan and King, Sharese},
  journal={Nature},
  volume={633},
  number={8028},
  pages={147--154},
  year={2024},
  publisher={Nature Publishing Group UK London}
}

@article{ji2023beavertails,
  title={Beavertails: Towards improved safety alignment of llm via a human-preference dataset},
  author={Ji, Jiaming and Liu, Mickel and Dai, Josef and Pan, Xuehai and Zhang, Chi and Bian, Ce and Chen, Boyuan and Sun, Ruiyang and Wang, Yizhou and Yang, Yaodong},
  journal={Advances in Neural Information Processing Systems},
  volume={36},
  pages={24678--24704},
  year={2023}
}

@article{liu2025towards,
  title={Towards realistic evaluation of cultural value alignment in large language models: Diversity enhancement for survey response simulation},
  author={Liu, Haijiang and Cao, Yong and Wu, Xun and Qiu, Chen and Gu, Jinguang and Liu, Maofu and Hershcovich, Daniel},
  journal={Information Processing \& Management},
  volume={62},
  number={4},
  pages={104099},
  year={2025},
  publisher={Elsevier}
}

@article{zhou2025fair,
  title={Fair-PP: A Synthetic Dataset for Aligning LLM with Personalized Preferences of Social Equity},
  author={Zhou, Qi and Zhang, Jie and Wang, Dongxia and Liu, Qiang and Li, Tianlin and Dong, Jin Song and Wang, Wenhai and Guo, Qing},
  journal={arXiv preprint arXiv:2505.11861},
  year={2025}
}

@misc{esomar2024global,
  author       = {{ESOMAR}},
  title        = {Global Market Research},
  year         = {2024},
  institution  = {ESOMAR}
}

@inproceedings{Kirk2024ThePA,
  title={The PRISM Alignment Dataset: What Participatory, Representative and Individualised Human Feedback Reveals About the Subjective and Multicultural Alignment of Large Language Models},
  author={Hannah Rose Kirk and Alexander Whitefield and Paul Rottger and Andrew M. Bean and Katerina Margatina and Juan Ciro and Rafael Mosquera and Max Bartolo and Adina Williams and He He and Bertie Vidgen and Scott A. Hale},
  booktitle={Neural Information Processing Systems},
  year={2024},
  url={https://api.semanticscholar.org/CorpusID:269362843}
}

@article{binz2024centaur,
  title={Centaur: a foundation model of human cognition},
  author={Binz, Marcel and Akata, Elif and Bethge, Matthias and Br{\"a}ndle, Franziska and Callaway, Fred and Coda-Forno, Julian and Dayan, Peter and Demircan, Can and Eckstein, Maria K and {\'E}ltet{\H{o}}, No{\'e}mi and others},
  journal={arXiv preprint arXiv:2410.20268},
  year={2024}
}

@article{cui2025large,
  title={A large-scale replication of scenario-based experiments in psychology and management using large language models},
  author={Cui, Ziyan and Li, Ning and Zhou, Huaikang},
  journal={Nature Computational Science},
  pages={1--8},
  year={2025},
  publisher={Nature Publishing Group US New York}
}

@article{Abdurahman2024PerilsAO,
  title={Perils and opportunities in using large language models in psychological research},
  author={Suhaib Abdurahman and Mohammad Atari and Farzan Karimi-Malekabadi and Mona J Xue and Jackson Trager and Peter S. Park and Preni Golazizian and Ali Omrani and Morteza Dehghani},
  journal={PNAS Nexus},
  year={2024},
  volume={3},
  url={https://api.semanticscholar.org/CorpusID:271240380}
}

@article{dominguez2024questioning,
  title={Questioning the Survey Responses of Large Language Models},
  author={Dominguez-Olmedo, Ricardo and Hardt, Moritz and Mendler-D{\"u}nner, Celestine},
  journal={Advances in Neural Information Processing Systems},
  year={2024}
}

@article{Kim2024FewshotPO,
  title={Few-shot Personalization of LLMs with Mis-aligned Responses},
  author={Jaehyung Kim and Yiming Yang},
  journal={ArXiv},
  year={2024},
  volume={abs/2406.18678},
  url={https://api.semanticscholar.org/CorpusID:270764657}
}

@article{Sun2024PersonaDBEL,
  title={Persona-DB: Efficient Large Language Model Personalization for Response Prediction with Collaborative Data Refinement},
  author={Chenkai Sun and Ke Yang and Revanth Gangi Reddy and Yi Ren Fung and Hou Pong Chan and ChengXiang Zhai and Heng Ji},
  journal={ArXiv},
  year={2024},
  volume={abs/2402.11060},
  url={https://api.semanticscholar.org/CorpusID:267750999}
}

@article{Chu2023LanguageMT,
  title={Language Models Trained on Media Diets Can Predict Public Opinion},
  author={Eric Chu and Jacob Andreas and Stephen Ansolabehere and Dwaipayan Roy},
  journal={ArXiv},
  year={2023},
  volume={abs/2303.16779},
  url={https://api.semanticscholar.org/CorpusID:257805183}
}

@inproceedings{He2024CommunityCrossInstructUI,
  title={Community-Cross-Instruct: Unsupervised Instruction Generation for Aligning Large Language Models to Online Communities},
  author={Zihao He and Rebecca Dorn and Siyi Guo and Minh Duc Hoang Chu and Kristina Lerman},
  booktitle={Conference on Empirical Methods in Natural Language Processing},
  year={2024},
  url={https://api.semanticscholar.org/CorpusID:270562726}
}

@article{Kwon2023EfficientMM,
  title={Efficient Memory Management for Large Language Model Serving with PagedAttention},
  author={Woosuk Kwon and Zhuohan Li and Siyuan Zhuang and Ying Sheng and Lianmin Zheng and Cody Hao Yu and Joseph E. Gonzalez and Haotong Zhang and Ion Stoica},
  journal={Proceedings of the 29th Symposium on Operating Systems Principles},
  year={2023},
  url={https://api.semanticscholar.org/CorpusID:261697361}
}

@article{Zhao2023GroupPO,
  title={Group Preference Optimization: Few-Shot Alignment of Large Language Models},
  author={Siyan Zhao and John Dang and Aditya Grover},
  journal={ArXiv},
  year={2023},
  volume={abs/2310.11523},
  url={https://api.semanticscholar.org/CorpusID:264289064}
}

@article{Lambert2024RewardBenchER,
  title={RewardBench: Evaluating Reward Models for Language Modeling},
  author={Nathan Lambert and Valentina Pyatkin and Jacob Daniel Morrison and Lester James Validad Miranda and Bill Yuchen Lin and Khyathi Raghavi Chandu and Nouha Dziri and Sachin Kumar and Tom Zick and Yejin Choi and Noah A. Smith and Hanna Hajishirzi},
  journal={ArXiv},
  year={2024},
  volume={abs/2403.13787},
  url={https://api.semanticscholar.org/CorpusID:268537409}
}

@article{Ethayarajh2024KTOMA,
  title={KTO: Model Alignment as Prospect Theoretic Optimization},
  author={Kawin Ethayarajh and Winnie Xu and Niklas Muennighoff and Dan Jurafsky and Douwe Kiela},
  journal={ArXiv},
  year={2024},
  volume={abs/2402.01306},
  url={https://api.semanticscholar.org/CorpusID:267406810}
}

@article{Kopf2023OpenAssistantC,
  title={OpenAssistant Conversations - Democratizing Large Language Model Alignment},
  author={Andreas Kopf and Yannic Kilcher and Dimitri von Rutte and Sotiris Anagnostidis and Zhi Rui Tam and Keith Stevens and Abdullah Barhoum and Nguyen Minh Duc and Oliver Stanley and Rich'ard Nagyfi and ES Shahul and Sameer Suri and David Glushkov and Arnav Varma Dantuluri and Andrew Maguire and Christoph Schuhmann and Huu Nguyen and Alexander Mattick},
  journal={ArXiv},
  year={2023},
  volume={abs/2304.07327},
  url={https://api.semanticscholar.org/CorpusID:258179434}
}

@article{Aroyo2023DICESDD,
  title={DICES Dataset: Diversity in Conversational AI Evaluation for Safety},
  author={Lora Aroyo and Alex S. Taylor and Mark D{\'i}az and Christopher Michael Homan and Alicia Parrish and Greg Serapio-Garc{\'i}a and Vinodkumar Prabhakaran and Ding Wang},
  journal={ArXiv},
  year={2023},
  volume={abs/2306.11247},
  url={https://api.semanticscholar.org/CorpusID:259203842}
}

@article{Binz2023TurningLL,
  title={Turning large language models into cognitive models},
  author={Marcel Binz and Eric Schulz},
  journal={ArXiv},
  year={2023},
  volume={abs/2306.03917},
  url={https://api.semanticscholar.org/CorpusID:259095948}
}

@article{tsai2025challenges,
  title={Challenges in adapting a survey: ensuring cross-cultural equivalence},
  author={Tsai, Tuan-I and Luck, Lauretta and Jefferies, Diana and Wilkes, Lesley},
  journal={Nurse researcher},
  volume={33},
  number={2},
  year={2025},
  publisher={RCN Publishing Company Limited}
}

@article{Santurkar2023WhoseOD,
  title={Whose Opinions Do Language Models Reflect?},
  author={Shibani Santurkar and Esin Durmus and Faisal Ladhak and Cinoo Lee and Percy Liang and Tatsunori Hashimoto},
  journal={ArXiv},
  year={2023},
  volume={abs/2303.17548},
  url={https://api.semanticscholar.org/CorpusID:257834040}
}

@article{loh2022youtube,
  title={Youtube dataset on mobile streaming for internet traffic modeling and streaming analysis},
  author={Loh, Frank and Wamser, Florian and Poign{\'e}e, Fabian and Gei{\ss}ler, Stefan and Ho{\ss}feld, Tobias},
  journal={Scientific Data},
  volume={9},
  number={1},
  pages={293},
  year={2022},
  publisher={Nature Publishing Group UK London}
}

@inproceedings{real2017youtube,
  title={Youtube-boundingboxes: A large high-precision human-annotated data set for object detection in video},
  author={Real, Esteban and Shlens, Jonathon and Mazzocchi, Stefano and Pan, Xin and Vanhoucke, Vincent},
  booktitle={Proceedings of the IEEE Conference on Computer Vision and Pattern Recognition},
  pages={5296--5305},
  year={2017}
}

@article{albadi2022deradicalizing,
  title={Deradicalizing YouTube: characterization, detection, and personalization of religiously intolerant Arabic videos},
  author={Albadi, Nuha and Kurdi, Maram and Mishra, Shivakant},
  journal={Proceedings of the ACM on Human-Computer Interaction},
  volume={6},
  number={CSCW2},
  pages={1--25},
  year={2022},
  publisher={ACM New York, NY, USA}
}

\newpage

\section{A. Dataset Documentation}
\subsection{A.1 Social Foundation Corpus}

The \textbf{Social Foundation Corpus} serves as the broad‐coverage pre-training layer for AlignSurvey. It consists of two coordinated components:

\begin{enumerate}
  \item \textbf{Qualitative component:} 44,021 fully anonymised interview dialogues curated from public video platforms and published academic sources related to oral history.  
  Dialogues longer than 2,000 tokens were centre–truncated and a five-turn sliding context window was generated for next-utterance prediction.
  \item \textbf{Quantitative component:} 411,174 structured survey records drawn from four flagship probability surveys (ATP, ESS, CSS, CGSS), covering trust in institutions, inequality, religion, media use, family dynamics and civic engagement.  Response options were randomly re-labelled to minimise order bias. 
\end{enumerate}

Details of each component are provided below.

\begin{table}[htp]
    \centering
    \small
    \begin{tabular}{l|cc}
    \toprule
       \textbf{Data} & \textbf{Item} & \textbf{Size} \\
    \midrule
    American Trends Panel(ATP) & Quan. & 106,028\\
    European Social Survey(ESS) & Quan. & 98,303\\
    Chinese Social Survey(CSS) & Quan. & 104,048\\
    Chinese General Social Survey(CGSS) & Quan. & 102,795\\
    Public Accessible Video Platforms & Qual. & 40,077\\
    Textual Excerpts from Academic Sources & Qual. & 3,944 \\
    \bottomrule
    \end{tabular}
    \caption{Origin, data modality (\textit{Quan.} = quantitative survey items; \textit{Qual.} = qualitative materials), and item count of the foundational datasets used in this study.}
    \label{tab:foundation_data}
\end{table}

\subsubsection{American Trends Panel (ATP):} 
Pew Research Center’s ATP is the only fully probability-based online panel that tracks the same U.S. adults over time.  Households without internet receive a tablet and mobile data plan, ensuring true national coverage.  We selected five waves with strong economic-attitudes overlap with our benchmark tasks.
These waves were chosen based on their relevance to key social survey themes used in dataset construction. The selected waves are listed in Table~\ref{tab:atp_wave}.

\begin{table}[htp]
    \centering
    \small
    \begin{tabular}{clc}
    \toprule
        \textbf{Wave} & \textbf{Wave Top} & \textbf{Population} \\
        \midrule
         40 & Trust, facts, and democracy & 10,618 \\
         41 & Views of America in 2050 & 2,526 \\
         54 & Economic inequality & 6,878 \\
         81 & Economics, pandemic financial outlook & 10,334\\
         103 & Economic well-being & 9,388 \\
         \bottomrule
    \end{tabular}
    \caption{ATP Survey Waves. Wave number, thematic focus, and sample size of the five American Trends Panel waves included in the quantitative tier of the Social Foundational Corpus.}
    \label{tab:atp_wave}
\end{table}

\subsubsection{European Social Survey (ESS):}
Since 2002 the ESS has fielded biennial, face-to-face probability surveys that apply identical sampling rules, translated instruments and centralised data processing across participating nations. Rigorous design—random dwelling selection, strict interviewer monitoring and double back-translation—makes the ESS a gold standard for cross-national comparisons of attitudes, beliefs and behavioural patterns. 
We selected Round 11 of the ESS for dataset construction. It is the most recent publicly available wave, covering 28 countries, and includes a rotating module on \emph{Justice \& Fairness} in addition to the core questionnaire battery. 

\subsubsection{Chinese Social Survey (CSS):}
Launched by the Institute of Sociology, Chinese Academy of Social Sciences in 2005, the CSS conducts biennial multi-stage, stratified probability interviews covering all 31 provincial-level units. By retaining a core module and rotating topical modules, CSS yields both time-series indicators of social change and flexibility to explore emerging issues in a rapidly transforming society. We selected the latest data from 2021 to build the dataset. Wave 2021 emphasised social mobility, government service quality and subjective class identity.

\subsubsection{Chinese General Social Survey (CGSS):}
Jointly organized by Renmin University of China and other institutions since 2003, the Chinese General Social Survey (CGSS) is the longest-running, nationally representative repeated cross-sectional survey of Chinese adults aged 18 and above. Its questionnaire combines fixed trend items with rotating modules covering topics such as social mobility, neighborhood life, political culture, and health. We selected the most recent wave from 2023 for dataset construction.


\subsubsection{Publicly Accessible Video Platforms:}

To simulate the conversational structure and thematic richness of in-person qualitative fieldwork, we collected 388 long-form interview videos on social issues from publicly accessible video platforms. Each recording was segmented into speaker turns (diarized), transcribed, and filtered for offensive language. All personal identifiers were anonymized prior to analysis. A full description of these preprocessing steps is provided in Appendix~A.3.


\subsubsection{Textual Excerpts from Academic Sources:}
We included textual materials from published works on qualitative social surveys and oral history, extracting interviewer–respondent exchanges from the text. These materials were processed using the same pipeline as the video interviews, including speaker segmentation, transcription, and anonymization to ensure appropriateness for academic use.


\begin{table}[htp]
    \centering
    \small
    \begin{tabular}{l|cc}
    \toprule
       \textbf{Data} & \textbf{Item}\\
    \midrule
       AlignSurvey-Expert(ASE) & Qual.+ Quan.\\
       General Social Survey(GSS) & Quan.\\
       China Household Income Project(CHIP) & Quan.\\
    \bottomrule
    \end{tabular}
    \caption{Entire-Pipeline Survey Datasets: overview of the three corpora that supervise the full modelling pipeline, with their respective data modalities (\textit{Qual.} = qualitative; \textit{Quan.} = quantitative).}
    \label{tab:entire_pipeline_overview}
\end{table}

\subsection{A.2 Entire-Pipeline Survey Datasets}

The \textbf{Entire-Pipeline Survey Datasets} supply explicit supervision for each of the four modelling stages—Social-Role Modelling, Semi-Structured-Interview Modelling, Attitude–Stance Modelling, and Survey-Response Modelling.  
An overview of all component corpora is given in Table \ref{tab:entire_pipeline_overview}, while corpus-level statistics for the Attitude, Stance, and Engagement (ASE) corpus appear in Table 3 of the main paper.
The demographic profile collected for every respondent is summarised in Table \ref{tab:demographic_ase}.

\paragraph{Qualitative component.}
The qualitative part of ASE comprises semi-structured interviews that cover eight broad themes (Table \ref{tab:inter_theme}).  

\begin{table*}[htp]
\centering
\begin{tabular}{l r l}
\hline
\textbf{Gender} & \textbf{Count} & \textbf{Percent} \\
\hline
Female & 3239 & \percbar{53.6}~53.6\% \\
Male & 2804 & \percbar{46.4}~46.4\% \\
\hline

\textbf{Age} & & \\
18--25 & 1945 & \percbar{32.2}~32.2\% \\
26--35 & 2151 & \percbar{35.6}~35.6\% \\
36--45 & 1103 & \percbar{18.3}~18.3\% \\
46--55 & 619 & \percbar{10.2}~10.2\% \\
56--65 & 192 & \percbar{3.2}~3.2\% \\
66--75 & 21 & \percbar{0.3}~0.3\% \\
$\geq 76$ & 12 & \percbar{0.2}~0.2\% \\
\hline

\textbf{Region} & & \\
Urban & 3300 & \percbar{54.6}~54.6\% \\
Rural & 2743 & \percbar{45.4}~45.4\% \\
\hline

\textbf{Occupation} & & \\
Sales & 1724 & \percbar{28.5}~28.5\% \\
Professional & 979 & \percbar{16.2}~16.2\% \\
Other & 968 & \percbar{16.0}~16.0\% \\
Student & 871 & \percbar{14.4}~14.4\% \\
Government Employee & 686 & \percbar{11.4}~11.4\% \\
Manual Labor & 482 & \percbar{8.0}~8.0\% \\
Retired & 160 & \percbar{2.6}~2.6\% \\
Business Owner & 151 & \percbar{2.5}~2.5\% \\
Self Employed & 10 & \percbar{0.2}~0.2\% \\
Company Employee & 8 & \percbar{0.1}~0.1\% \\
Unemployed & 4 & \percbar{0.1}~0.1\% \\
\hline

\textbf{Education} & & \\
Bachelor's Degree & 3365 & \percbar{55.7}~55.7\% \\
Associate Degree & 1177 & \percbar{19.4}~19.4\% \\
Graduate Degree & 800 & \percbar{13.3}~13.3\% \\
High School & 556 & \percbar{9.2}~9.2\% \\
Junior High School & 145 & \percbar{2.4}~2.4\% \\
\hline

\textbf{Home Ownership} & & \\
Yes & 3112 & \percbar{51.5}~51.5\% \\
No & 2931 & \percbar{48.5}~48.5\% \\
\hline

\textbf{Car Ownership} & & \\
Yes & 3180 & \percbar{52.6}~52.6\% \\
No & 2863 & \percbar{47.4}~47.4\% \\
\hline

\textbf{Household Size} & & \\
0 & 1 & \percbar{0.0}~0.0\% \\
1 & 62 & \percbar{1.0}~1.0\% \\
2 & 669 & \percbar{11.1}~11.1\% \\
3 & 2219 & \percbar{36.7}~36.7\% \\
4 & 1453 & \percbar{24.0}~24.0\% \\
5 & 959 & \percbar{15.9}~15.9\% \\
6 & 427 & \percbar{7.1}~7.1\% \\
7 & 126 & \percbar{2.1}~2.1\% \\
$\geq 8$ & 127 & \percbar{2.1}~2.1\% \\
\hline

\textbf{Income} & & \\
110k--200k & 1985 & \percbar{32.8}~32.8\% \\
210k--500k & 1748 & \percbar{28.9}~28.9\% \\
60k--100k & 1179 & \percbar{19.5}~19.5\% \\
$\leq 50k$ & 647 & \percbar{10.7}~10.7\% \\
600k$\geq$ & 263 & \percbar{4.4}~4.4\% \\
510k--600k & 221 & \percbar{3.7}~3.7\% \\
\hline

\textbf{Stress Level} & & \\
Moderate Stress & 2695 & \percbar{44.6}~44.6\% \\
Mild Stress & 1898 & \percbar{31.4}~31.4\% \\
High Stress & 911 & \percbar{15.1}~15.1\% \\
No Stress & 351 & \percbar{5.8}~5.8\% \\
Very High Stress & 188 & \percbar{3.1}~3.1\% \\
\hline
\end{tabular}
\caption{Task4 Demographic Distribution with Visual Percent Bars}
\label{tab:demographics}
\end{table*}

\paragraph{Quantitative component.}
The quantitative part of ASE is a county-level development-and-governance questionnaire.  
Its hierarchical theme structure—parent topics and corresponding survey questions—is shown in Table \ref{tab:survey_themes}.  

\begin{table}[htp]
    \centering
    \begin{tabular}{l|l}
    \toprule
        \textbf{ID} & \textbf{Theme} \\
    \midrule
         1 & Policy understanding \\
         2 & Policy details known \\
         3 & Policy awareness \\
         4 & Income level awareness \\
         5 & Public service evaluation \\
         6 & Social class attitude \\
         7 & Upward mobility attitude \\
         8 & Future policy expectations \\
    \bottomrule
    \end{tabular}
    \caption{Interview themes in the ASE qualitative dataset.}
    \label{tab:inter_theme}
\end{table}

\begin{table*}[htp]
  \centering
  \renewcommand{\arraystretch}{1.2}   
  \setlength{\tabcolsep}{4pt}         
  \begin{tabular}{%
    >{\raggedright\arraybackslash}p{0.36\linewidth}|
    >{\raggedright\arraybackslash}p{0.64\linewidth}}
    \toprule
    \textbf{Attribute} & \textbf{Option Value} \\
    \midrule
    Gender & 
      Male; Female \\[2pt]

    Age & 
      18–25 years; 26–35 years; 36–45 years; 46–55 years; 56–65 years; 66–75 years; 76 years and above \\[2pt]

    Region & 
      Urban; Rural \\[2pt]

    Education & 
      Primary School; Junior High School; High School; Associate Degree; Bachelor’s Degree; Graduate Degree \\[2pt]

    Home Ownership & 
      Yes; No \\[2pt]

    Car Ownership & 
      Yes; No \\[2pt]

    Household Income & 
      Below ¥50,000; ¥50,000–200,000; ¥200,000–500,000; ¥500,000–1,000,000; Above ¥1,000,000 \\[2pt]

    Stress Level & 
      No Stress; Mild Stress; Moderate Stress; High Stress; Very High Stress \\[2pt]

    Occupation & 
      Student; Retired; Company Employee; Government Employee; Private Business Owner; Self-employed Merchant; Other \\[2pt]

    Registered Residence & 
     Geographical location \\[2pt]

    Current Residence & 
      Geographical location \\[2pt]

    Household Size & 
      Integer, e.g., 4 \\[2pt]

    Family Relations & 
      e.g., spouse, children \\[2pt]

    Main Monthly Expenses & 
      e.g., mortgage, education, healthcare \\[2pt]
    \bottomrule
  \end{tabular}
  \caption{Demographic attributes collected for participants in the ASE.}
  \label{tab:demographic_ase}
\end{table*}

\begin{table*}[htp]
\centering
\begin{tabular}{l|l}
\toprule
\textbf{ID} & \textbf{Survey Question} \\
\midrule
\multicolumn{2}{l}{\textbf{Economic Development}}\\
1.1  & Perception that the county’s economy is developing rapidly \\
1.2 & Perception that the county’s economic‐development model is sustainable \\
1.3 & Improvement in local job opportunities and salary levels \\
1.4 & Government effort to promote high-quality economic development \\
\midrule
\multicolumn{2}{l}{\textbf{Social Gaps and Fairness}}\\
2.1 & Evaluation of the income gap \\
2.2 & Evaluation of the urban–rural gap \\
2.3 & Evaluation of the regional gap \\
2.4 & Perception that local society is fair \\
\midrule
\multicolumn{2}{l}{\textbf{Public-Service Satisfaction}}\\
3.1  & Satisfaction with public education services \\
3.2  & Satisfaction with public medical services \\
3.3  & Satisfaction with public elderly-care services \\
3.4  & Satisfaction with public childcare services \\
3.5  & Satisfaction with public employment services \\
3.6  & Satisfaction with public housing services \\
3.7  & Satisfaction with public assistance services \\
3.8  & Satisfaction with public transportation services \\
\midrule
\multicolumn{2}{l}{\textbf{Governance and Participation}}\\
4.1 & Perception that the social environment is safe and stable \\
4.2  & Perception that the county’s governance level has improved \\
4.3  & Ability to participate in community or village affairs \\
4.4  & Satisfaction with the current government’s development performance \\
4.5  & Confidence in the government’s future development \\
\bottomrule
\end{tabular}
\caption{Hierarchical topic structure of the ASE quantitative questionnaire: parent themes and their child survey questions}
\label{tab:survey_themes}
\end{table*}

\subsection{A.3 Pre-processing and Splits}

All interview recordings in the qualitative datasets are first transcribed with an automatic speech-recognition system; subsequently, human editors sample and review 2 \% of the material to bring the transcripts to near-perfect orthographic accuracy, intervening in disfluencies only when these obstruct comprehension. After transcription, we run a conservative, rule-based anonymiser that leverages GPT-4o: every overt personal identifier—names, institutions, addresses, and similar noun phrases—is replaced with a unique placeholder (e.g., “I, $<NAME\_17C>$, was admitted to Harvard University in 1998”). Each interview alternates between interviewer (A) and interviewee (B), though turn lengths vary. We tokenise sessions into speaker-aligned turns and apply a five-turn sliding window that always begins with interviewer A and advances by one turn at a time (e.g., for A B A B A B A, window 1 spans turns 1–5, window 2 spans turns 3–7, etc.). Windows are discarded if their concatenated text, after trimming whitespace, contains fewer than 20 alphabetic tokens or if anonymisation has redacted more than 30 \% of the tokens.

In the quantitative datasets, every single-choice answer is shuffled at load time, as described in the main paper. To standardise rating scales, Likert inventories with an odd number of anchors are remapped onto a canonical three-class backbone: edge bins are collapsed in proportion to each item’s empirical distribution so that the resulting histogram is as balanced as possible. Scales with an even number of anchors, which lack a neutral midpoint by design, are instead projected onto a four-class backbone. To prevent the model from memorising positional biases, we also shuffle the letters associated with each option as well as the option texts themselves.

For the Social Foundation Corpus, all datasets are ingested to fine-tune the backbone model via supervised instruction (SFT), thereby imparting foundational knowledge before evaluation. Test splits of the Entire-Pipeline Survey Datasets are designed to probe generalisation across both demographic profiles and thematic domains. Each respondent receives a fingerprint composed of six categorical pillars—gender, age, region status, highest educational level, household-income bracket, and occupation category. Respondents whose fingerprint never appears in the provisional training pool are assigned to the hold-out group, and they constitute exactly half of the final test set; the remaining half consists of respondents who do occur in training but answer different questions or segments, enforcing cross-topic rather than cold-start extrapolation. Concretely, the ASE split is reported in Table 3 of the main paper; the GSS split contains 18 922 training and 4 730 test instances, while the CHIP split comprises 11 718 training and 3 129 test instances.
\section{B. Comprehensive Experimental Results}

\subsection{B.1 Task1 Social Role Modeling}
Table \ref{tab:demo_socio_landscape} presents detailed results for Task 1: Social Role Modeling. This task aims to evaluate the capability of various large language models (LLMs) to infer demographic and socioeconomic attributes based solely on provided dialogues.

The SurveyLM series significantly outperform general-purpose LLMs across all categories (Basic INFO, Social INFO, and Family INFO). Notably, SurveyLM$_{\text{Qwen2.5-7B}}$ achieves the highest overall accuracy, especially excelling in predicting "Gender" (67.81\%), "Age" (50.62\%), and "Income" (40.42\%). General models like GPT-4o and Claude show a moderate ability to predict demographic traits, but their performance drops dramatically in complex categories like "Relationship" and "Expenses."

Few-shot prompts improve the performance of general models, such as GPT-4o (from 0.00\% to 23.75\% in "Relationship"). However, even with few-shot learning, these models remain significantly below SurveyLM performance. Models with smaller scales generally show weaker performance, particularly in zero-shot settings, underscoring the importance of specialized fine-tuning for accurate demographic inference.

\begin{sidewaystable}[htp]
  \centering
  \small
  \begin{tabular}{l|ccccc|cccc|ccc}
    \toprule
    \multicolumn{1}{c|}{\textbf{Model}} &
    \multicolumn{5}{c|}{\textbf{Basic INFO}} &
    \multicolumn{4}{c|}{\textbf{Social INFO}} &
    \multicolumn{3}{c}{\textbf{Family INFO}} \\
    \cmidrule(lr){2-6}\cmidrule(lr){7-10}\cmidrule(lr){11-13}
     & \textbf{Gender} & \textbf{Age} & \textbf{Area} & \textbf{Edu.} & \textbf{Employ.} &
       \textbf{Stress} & \textbf{Home Own.} & \textbf{Veh. Own.} & \textbf{Income} &
       \textbf{Rel.} & \textbf{HH Size} & \textbf{Expenses} \\ 
    \midrule
    GPT-4o \textit{(Zero-shot)}     & 50.00\% & 37.08\% & 53.75\% & 34.17\% & 45.42\% &  9.58\% & 47.92\% & 48.33\% & 11.67\% &  0.00\% &  3.75\% &  0.00\% \\
    GPT-4o \textit{(Few-shot)}      & 52.25\% & 40.25\% & 39.50\% & 34.75\% & 36.20\% & 11.80\% & 48.10\% & 49.85\% & 19.22\% & 23.75\% &  2.50\% & 24.25\% \\
    Claude \textit{(Zero-shot)}     & 51.92\% & 35.00\% & 57.92\% & 23.75\% & 45.00\% & 12.08\% & 47.50\% & 42.92\% &  7.50\% &  0.00\% &  7.92\% &  0.83\% \\
    Claude \textit{(Few-shot)}      & 51.10\% & 40.10\% & 37.05\% & 39.10\% & 33.40\% & 11.40\% & 44.35\% & 43.80\% & 11.13\% & 21.25\% &  0.40\% & 24.60\% \\
    DeepSeek\textminus R1 \textit{(Zero-shot)} & 51.33\% & 34.58\% & 55.83\% & 12.50\% & 38.33\% &  1.67\% & 52.08\% & 49.17\% &  8.75\% &  0.00\% & 15.83\% &  1.67\% \\
    DeepSeek\textminus R1 \textit{(Few-shot)}  & 51.05\% & 40.00\% & 42.60\% & 31.20\% & 16.95\% &  2.45\% & 54.55\% & 34.60\% & 11.67\% & 20.60\% & 17.65\% & 23.15\% \\
    R1\textminus Distill\textminus Qwen\textminus 14B \textit{(Zero-shot)} & 38.75\% & 21.67\% & 55.00\% & 14.17\% & 21.67\% &  0.00\% & 50.42\% & 30.83\% &  0.00\% &  0.00\% & 11.25\% &  0.00\% \\
    R1\textminus Distill\textminus Qwen\textminus 14B \textit{(Few-shot)}  & 40.60\% & 43.60\% & 42.25\% & 24.05\% & 23.10\% &  2.55\% & 52.45\% & 38.55\% &  8.11\% & 20.30\% & 12.15\% & 22.90\% \\
    Qwen2.5\textminus 72B \textit{(Zero-shot)} & 48.20\% & 43.95\% & 50.51\% & 17.27\% & 29.24\% &  1.23\% & 48.77\% & 50.00\% &  4.27\% &  0.00\% &  8.22\% &  0.00\% \\
    Qwen2.5\textminus 72B \textit{(Few-shot)}  & 46.05\% & 43.30\% & 47.55\% & 23.25\% & 33.10\% &  1.50\% & 60.95\% & 35.80\% &  9.67\% & 20.35\% & 14.60\% & 23.00\% \\
    \midrule
    Meta\textminus Llama\textminus 3\_1\textminus 8B \textit{(Zero-shot)} & 47.50\% & 40.42\% & 58.33\% & 12.50\% & 26.25\% &  0.00\% & 50.83\% & 50.83\% &  6.67\% &  0.00\% & 26.67\% &  0.00\% \\
    Meta\textminus Llama\textminus 3\_1\textminus 8B \textit{(Few-shot)}  & 49.05\% & 40.05\% & 41.55\% & 23.10\% & 23.35\% &  4.40\% & 47.10\% & 46.05\% &  9.11\% & 20.45\% & 11.55\% & 22.95\% \\
    Mistral\textminus 7B\textminus v0.3 \textit{(Zero-shot)} & 39.17\% & 35.00\% & 48.75\% &  4.58\% & 35.42\% &  3.33\% & 51.67\% & 50.42\% & 19.17\% &  0.00\% &  0.83\% &  0.42\% \\
    Mistral\textminus 7B\textminus v0.3 \textit{(Few-shot)}  & 22.80\% & 43.10\% & 43.35\% & 55.05\% & 32.25\% &  6.35\% & 52.60\% & 39.25\% & 13.33\% & 20.40\% &  9.45\% & 22.95\% \\
    Qwen2.5\textminus 7B \textit{(Zero-shot)} & 46.67\% & 41.67\% & 49.58\% & 15.83\% & 26.67\% &  0.83\% & 50.83\% & 50.00\% &  4.17\% &  0.00\% &  6.67\% &  0.00\% \\
    Qwen2.5\textminus 7B \textit{(Few-shot)}  & 48.55\% & 43.45\% & 45.10\% & 17.25\% & 35.25\% &  1.65\% & 60.10\% & 51.75\% &  9.83\% & 20.55\% & 13.70\% & 22.95\% \\
    \midrule
    $SurveyLM_{\text{Mistral-7B-v0.3}}$       & 54.31\% & 45.87\% & 59.67\% & 37.90\% & 43.47\% & 35.44\% & 60.54\% & 60.71\% & 33.18\% & 85.88\% & 39.99\% & 29.28\% \\
    $SurveyLM_{\text{Meta-Llama-3\_1-8B}}$    & 55.14\% & 47.03\% & 62.05\% & 41.02\% & 47.49\% & 44.02\% & 65.92\% & 65.01\% & 31.29\% & 80.06\% & 34.80\% & 32.78\% \\
    $SurveyLM_{\text{Qwen2.5-7B}}$            & 67.81\% & 50.62\% & 60.58\% & 41.83\% & 52.43\% & 36.55\% & 68.25\% & 66.88\% & 40.42\% & 86.42\% & 34.04\% & 33.95\% \\
    \bottomrule
  \end{tabular}
  \caption{Task1 Social Role Modeling Comprehensive Results.}
  \label{tab:demo_socio_landscape}
\end{sidewaystable}

\subsection{B.2 Task2 Semi-Structured Interview Modeling}

Table \ref{tab:conv_eval} presents evaluation scores on naturalness, style matching, and consistency of generated interview responses.

SurveyLM models exhibit significantly better consistency (around 3.00) and naturalness (3.77 to 3.98) compared to general-purpose LLMs (GPT-4o, Claude 3.7 Sonnet). Particularly, SurveyLM$_{\text{Qwen2.5-7B}}$ achieves the highest score for "Naturalness" (3.98) and "Consistency" (3.01), reflecting effective alignment with the semi-structured interview style. 
General models, while showing acceptable "Naturalness" (around 3.40 to 3.49), fall short in "Style Match" (below 3.00), indicating limitations in adapting their responses to the specific dialogue style required in semi-structured interviews.
Fine-tuning substantially enhances the capability of LLMs to produce coherent, contextually appropriate interview responses, emphasizing the necessity for task-specific fine-tuning strategies in qualitative modeling tasks.

\begin{table*}[htp]
  \centering
  \begin{tabular}{l|ccc}
    \toprule
    \multicolumn{1}{c|}{\textbf{Model}} &
    \textbf{Naturalness} $\uparrow$ &
    \textbf{Style Match} $\uparrow$ &
    \textbf{Consistency} $\uparrow$ \\
    \midrule
    GPT-4o \textit{(Zero-shot)} & 3.45 & 2.85 & 2.89 \\
    GPT-4o \textit{(Few-shot)}  & 3.49 & 2.80 & 2.75 \\
    Claude 3.7 Sunnet \textit{(Zero-shot)} & 3.41 & 2.81 & 2.85 \\
    Claude 3.7 Sunnet \textit{(Few-shot)}  & 3.48 & 2.83 & 2.90 \\
    DeepSeek-R1 \textit{(Zero-shot)} & 3.40 & 2.81 & 2.83 \\
    DeepSeek-R1 \textit{(Few-shot)}  & 3.47 & 2.70 & 2.78 \\
    R1-Distill-Qwen-14B \textit{(Zero-shot)} & 2.78 & 2.20 & 2.36 \\
    R1-Distill-Qwen-14B \textit{(Few-shot)}  & 2.85 & 2.25 & 2.30 \\
    Qwen2.5-72B \textit{(Zero-shot)} & 3.40 & 2.72 & 2.85 \\
    Qwen2.5-72B \textit{(Few-shot)}  & 3.46 & 2.65 & 2.70 \\
    \midrule
    Meta-Llama-3.1-8B \textit{(Zero-shot)} & 3.30 & 2.69 & 2.79 \\
    Meta-Llama-3.1-8B \textit{(Few-shot)}  & 3.36 & 2.60 & 2.68 \\
    Mistral-7B-v0.3 \textit{(Zero-shot)} & 3.21 & 2.57 & 2.44 \\
    Mistral-7B-v0.3 \textit{(Few-shot)}  & 3.28 & 2.50 & 2.40 \\
    Qwen2.5-7B \textit{(Zero-shot)} & 3.39 & 2.67 & 2.79 \\
    Qwen2.5-7B \textit{(Few-shot)}  & 3.44 & 2.60 & 2.85 \\
    \midrule
    $SurveyLM_{\text{Mistral-7B-v0.3}}$ & 2.61 & 2.90 & 2.90 \\
    $SurveyLM_{\text{Meta-Llama-3 1-8B}}$ & 3.77 & 3.00 & 2.94 \\
    $SurveyLM_{\text{Qwen2.5-7B}}$ & 3.98 & 2.96 & 3.01 \\
    \bottomrule
  \end{tabular}
  \caption{Task2 Semi-Structured Interview Modeling Result.}
  \label{tab:conv_eval}
\end{table*}

\subsection{B.3 Task3 Attitude Stance Modeling}

Table \ref{tab:auto_metrics_landscape} outlines detailed metrics evaluating attitude stance prediction.

SurveyLM models outperform general-purpose models on metrics such as Accuracy, Precision, Recall, and F1-score. Notably, SurveyLM$_{\text{Mistral-7B-v0.3}}$ attains the highest F1-score (53.85\%), showing strong balance across precision and recall. GPT-4o, despite its general capabilities, demonstrates relatively lower performance (F1 around 29.35\% to 32.78\%) indicating challenges in nuanced attitude stance recognition.
Models in zero-shot settings typically have lower scores, suggesting limited inherent capability to accurately model attitudes without explicit domain-specific training. The dramatic improvement in Jaccard similarity and Cosine similarity scores for SurveyLM models indicates enhanced quality of generated reasoning chains compared to general-purpose LLMs.
The substantial gains achieved by SurveyLM models across multiple metrics confirm that targeted fine-tuning significantly improves attitude stance modeling, enabling better representation of nuanced human attitudes.

\begin{table*}[htp]
  \centering
  \small
  \setlength{\tabcolsep}{4pt}
  \begin{tabular}{l|cccccccc}
    \toprule
    \textbf{Model} &
    \textbf{ACC} $\uparrow$ &
    \textbf{PRE} $\uparrow$ &
    \textbf{RECALL} $\uparrow$ &
    \textbf{F1} $\uparrow$ &
    \textbf{ROUGE} $\uparrow$ &
    \textbf{Jaccard} $\uparrow$ &
    \textbf{Cosine} $\uparrow$ &
    \textbf{WD} $\downarrow$ \\
    \midrule
    GPT-4o \textit{(Zero-shot)} & 38.12\% & 32.07\% & 32.41\% & 29.35\% & 9.83 & 0.0987 & 0.0125 & 1.6479 \\
    GPT-4o \textit{(Few-shot)} & 42.73\% & 35.46\% & 34.04\% & 32.78\% & 9.10 & 0.1031 & 0.0137 & 1.1800 \\
    Claude 3.7 Sonnet \textit{(Zero-shot)} & 28.12\% & 20.72\% & 17.71\% & 17.35\% & 8.16 & 0.0637 & 0.0002 & 1.3319 \\
    Claude 3.7 Sonnet \textit{(Few-shot)} & 31.48\% & 23.63\% & 20.97\% & 19.89\% & 8.10 & 0.0715 & 0.0006 & 1.1200 \\
    DeepSeek-R1 \textit{(Zero-shot)} & 36.88\% & 35.05\% & 33.50\% & 32.06\% & 8.19 & 0.0745 & 0.0026 & 1.3561 \\
    DeepSeek-R1 \textit{(Few-shot)} & 41.05\% & 36.27\% & 34.66\% & 33.86\% & 8.12 & 0.0810 & 0.0034 & 1.1000 \\
    R1-Distill-Qwen-14B \textit{(Zero-shot)} & 38.75\% & 35.09\% & 34.93\% & 34.36\% & 8.44 & 0.1045 & 0.0058 & 1.7952 \\
    R1-Distill-Qwen-14B \textit{(Few-shot)} & 33.81\% & 29.11\% & 28.46\% & 28.02\% & 8.20 & 0.1111 & 0.0076 & 1.1400 \\
    Qwen2.5-72B \textit{(Zero-shot)} & 35.00\% & 32.51\% & 31.04\% & 31.05\% & 9.95 & 0.1015 & 0.0097 & 1.6533 \\
    Qwen2.5-72B \textit{(Few-shot)} & 42.16\% & 38.92\% & 37.42\% & 36.90\% & 8.35 & 0.1126 & 0.0141 & 1.0500 \\
    \midrule
    Mistral-7B-v0.3 \textit{(Zero-shot)} & 38.44\% & 35.11\% & 33.98\% & 33.05\% & 8.35 & 0.1056 & 0.0121 & 1.5344 \\
    Mistral-7B-v0.3 \textit{(Few-shot)} & 39.77\% & 38.92\% & 35.86\% & 34.31\% & 7.55 & 0.1144 & 0.0152 & 0.3023 \\
    Meta-Llama-3.1-8B \textit{(Zero-shot)} & 39.38\% & 32.65\% & 31.99\% & 29.45\% & 8.59 & 0.1068 & 0.0078 & 1.5218 \\
    Meta-Llama-3.1-8B \textit{(Few-shot)} & 39.96\% & 38.26\% & 35.77\% & 34.91\% & 7.70 & 0.1185 & 0.0139 & 0.4470 \\
    Qwen2.5-7B \textit{(Zero-shot)} & 36.88\% & 38.20\% & 37.69\% & 29.78\% & 9.50 & 0.1040 & 0.0123 & 1.7000 \\
    Qwen2.5-7B \textit{(Few-shot)} & 44.69\% & 39.52\% & 38.51\% & 34.87\% & 7.60 & 0.1183 & 0.0148 & 0.3720 \\
    \midrule
    $SurveyLM_{\text{Mistral-7B-v0.3}}$    & 57.13\% & 54.44\% & 53.36\% & 53.85\% & 7.64 & 0.1308 & 0.0249 & 0.2974 \\
    $SurveyLM_{\text{Meta-Llama-3\_1-8B}}$ & 55.09\% & 54.77\% & 51.02\% & 52.14\% & 7.86 & 0.1341 & 0.0189 & 0.4583 \\
    $SurveyLM_{\text{Qwen2.5-7B}}$         & 56.83\% & 42.74\% & 41.49\% & 41.52\% & 7.69 & 0.1281 & 0.0172 & 0.3851 \\
    \bottomrule
  \end{tabular}
  \caption{Task3 Attitude Stance Modeling Comprehensive Result.}
  \label{tab:auto_metrics_landscape}
\end{table*}

\subsection{B.4 Task4 Structured Response Modeling}

Tables \ref{tab:oursurvey}, \ref{tab:chip}, and \ref{tab:gss} showcase results for structured response modeling across multiple datasets (ASE, CHIP, and GSS).
SurveyLM models demonstrate superior performance consistently across ASE, CHIP, and GSS datasets. Notably, SurveyLM$_{\text{Qwen2.5-7B}}$ achieves the lowest Wasserstein Distance (WD) indicating excellent distribution alignment with real-world responses. Compared to general-purpose models, SurveyLM notably improves Accuracy (around 51\% to 67\%) and F1 scores (34\% to 41\%), underscoring its robustness in modeling structured survey responses.
General-purpose models typically underperform significantly, especially in zero-shot settings (Accuracy below 50\%, F1 scores often below 25\%), revealing limited capabilities in structured question-answering tasks without specialized training.
These findings reinforce the critical role of targeted fine-tuning, as demonstrated by the SurveyLM series, in effectively modeling structured survey responses and achieving high fidelity and consistency with real-world human responses.

\begin{table*}[htp]
  \centering
  \begin{tabular}{l|ccccc}
    \toprule
    \multicolumn{1}{c|}{\textbf{Model}} &
    \textbf{ACC} $\uparrow$ &
    \textbf{PRE} $\uparrow$ &
    \textbf{RECALL} $\uparrow$ &
    \textbf{F1} $\uparrow$ &
    \textbf{WD} $\downarrow$ \\
    \midrule
    GPT\!-4o \textit{(Zero-shot)}    & 46.56 & 37.68 & 23.94 & 22.05 & 2.1336 \\
    GPT\!-4o \textit{(Few-shot)}     & 47.91 & 38.40 & 23.40 & 23.10 & 1.7895 \\
    Claude 3.7 Sonnet \textit{(Zero-shot)} & 42.11 & 21.98 & 22.91 & 18.44 & 1.7789 \\
    Claude 3.7 Sonnet \textit{(Few-shot)}  & 42.95 & 22.30 & 22.40 & 19.50 & 1.6542 \\
    DeepSeek-R1 \textit{(Zero-shot)} & 47.21 &  9.22 &  7.54 &  7.11 & 1.9987 \\
    DeepSeek-R1 \textit{(Few-shot)}  & 47.88 &  9.05 &  8.20 &  7.65 & 1.8527 \\
    R1-Distill-Qwen-14B \textit{(Zero-shot)} & 49.54 & 28.82 & 31.96 & 25.90 & 1.9295 \\
    R1-Distill-Qwen-14B \textit{(Few-shot)}  & 50.29 & 29.70 & 31.60 & 26.80 & 1.7164 \\
    Qwen2.5-72B \textit{(Zero-shot)} & 42.20 &  5.24 &  3.87 &  3.80 & 1.8654 \\
    Qwen2.5-72B \textit{(Few-shot)}  & 43.06 &  5.60 &  3.70 &  4.05 & 1.7348 \\
    \midrule
    Meta-Llama-3.1-8B \textit{(Zero-shot)} & 14.84 & 0.37 & 0.13 & 0.17 & 2.1154 \\
    Meta-Llama-3.1-8B \textit{(Few-shot)}  & 17.26 & 0.42 & 0.12 & 0.19 & 1.9451 \\
    Mistral-7B-v0.3 \textit{(Zero-shot)}   & 28.04 & 1.85 & 1.07 & 1.10 & 2.2118 \\
    Mistral-7B-v0.3 \textit{(Few-shot)}    & 30.47 & 2.10 & 0.95 & 1.25 & 2.0823 \\
    Qwen2.5-7B \textit{(Zero-shot)}        & 16.39 & 0.27 & 0.10 & 0.12 & 1.8479 \\
    Qwen2.5-7B \textit{(Few-shot)}         & 18.46 & 0.24 & 0.14 & 0.15 & 1.7059 \\
    \midrule
    $AlignSurvey_{\text{Mistral-7B-v0.3}}$     & 54.59 & 45.30 & 41.01 & 41.34 & 0.0671 \\
    $AlignSurvey_{\text{Meta-Llama-3.1-8B}}$   & 56.11 & 44.97 & 41.68 & 41.47 & 0.0675 \\
    $AlignSurvey_{\text{Qwen2.5-7B}}$          & 55.47 & 44.23 & 41.87 & 41.79 & 0.0541 \\
    \bottomrule
  \end{tabular}
  \caption{Task4 Structured Response Modeling Results on the \textsc{ASE}.}
  \label{tab:oursurvey}
\end{table*}

\begin{table*}[htp]
  \centering
  \begin{tabular}{l|ccccc}
    \toprule
    \multicolumn{1}{c|}{\textbf{Model}} &
    \textbf{ACC} $\uparrow$ &
    \textbf{PRE} $\uparrow$ &
    \textbf{RECALL} $\uparrow$ &
    \textbf{F1} $\uparrow$ &
    \textbf{WD} $\downarrow$ \\
    \midrule
    GPT\!-4o \textit{(Zero-shot)}    & 53.43 & 32.18 & 30.61 & 26.24 & 1.9325 \\
    GPT\!-4o \textit{(Few-shot)}     & 54.32 & 32.74 & 30.20 & 27.30 & 1.6992 \\
    Claude 3.7 Sonnet \textit{(Zero-shot)} & 47.21 & 12.50 & 10.54 &  9.50 & 1.8872 \\
    Claude 3.7 Sonnet \textit{(Few-shot)}  & 47.98 & 13.02 & 11.20 &  9.95 & 1.7318 \\
    DeepSeek-R1 \textit{(Zero-shot)} & 44.65 &  4.55 &  8.65 &  4.52 & 1.7789 \\
    DeepSeek-R1 \textit{(Few-shot)}  & 45.27 &  4.83 &  9.10 &  4.90 & 1.6523 \\
    R1-Distill-Qwen-14B \textit{(Zero-shot)} & 50.62 & 31.77 & 37.76 & 29.36 & 1.9448 \\
    R1-Distill-Qwen-14B \textit{(Few-shot)}  & 51.37 & 32.35 & 37.10 & 30.67 & 1.7104 \\
    Qwen2.5-72B \textit{(Zero-shot)} & 47.98 & 18.23 & 18.02 & 13.29 & 1.9918 \\
    Qwen2.5-72B \textit{(Few-shot)}  & 48.66 & 18.58 & 17.80 & 14.02 & 1.7989 \\
    \midrule
    Meta-Llama-3.1-8B \textit{(Zero-shot)} & 20.84 &  4.30 &  2.09 &  2.40 & 1.7748 \\
    Meta-Llama-3.1-8B \textit{(Few-shot)}  & 22.46 &  4.68 &  1.95 &  2.60 & 1.6435 \\
    Mistral-7B-v0.3 \textit{(Zero-shot)}   & 23.60 &  5.73 &  3.63 &  3.46 & 1.9941 \\
    Mistral-7B-v0.3 \textit{(Few-shot)}    & 26.89 &  6.12 &  3.40 &  3.70 & 1.8876 \\
    Qwen2.5-7B \textit{(Zero-shot)}        & 51.95 & 13.44 & 17.05 & 12.51 & 1.9099 \\
    Qwen2.5-7B \textit{(Few-shot)}         & 53.14 & 13.98 & 17.55 & 13.15 & 1.7441 \\
    \midrule
    $AlignSurvey_{\text{Mistral-7B-v0.3}}$     & 65.56 & 35.90 & 38.49 & 35.35 & 0.0857 \\
    $AlignSurvey_{\text{Meta-Llama-3.1-8B}}$   & 64.89 & 32.91 & 36.82 & 32.99 & 0.0845 \\
    $AlignSurvey_{\text{Qwen2.5-7B}}$          & 67.17 & 36.48 & 38.82 & 34.94 & 0.0686 \\
    \bottomrule
  \end{tabular}
  \caption{Task4 Structured Response Modeling Results on the \textsc{CHIP}.}
  \label{tab:chip}
\end{table*}

\begin{table*}[htp]
  \centering
  \begin{tabular}{l|ccccc}
    \toprule
    \multicolumn{1}{c|}{\textbf{Model}} &
    \textbf{ACC} $\uparrow$ &
    \textbf{PRE} $\uparrow$ &
    \textbf{RECALL} $\uparrow$ &
    \textbf{F1} $\uparrow$ &
    \textbf{WD} $\downarrow$ \\
    \midrule
    GPT\!-4o \textit{(Zero-shot)}    & 32.79 & 27.19 & 26.94 & 24.41 & 1.1130 \\
    GPT\!-4o \textit{(Few-shot)}     & 33.85 & 27.90 & 26.40 & 25.10 & 1.0284 \\
    Claude 3.7 Sonnet \textit{(Zero-shot)} & 33.94 & 30.72 & 28.41 & 25.93 & 1.2248 \\
    Claude 3.7 Sonnet \textit{(Few-shot)}  & 34.55 & 30.50 & 28.10 & 26.60 & 1.1095 \\
    DeepSeek-R1 \textit{(Zero-shot)} & 34.54 & 29.64 & 27.63 & 23.15 & 1.3854 \\
    DeepSeek-R1 \textit{(Few-shot)}  & 35.20 & 29.20 & 28.00 & 23.70 & 1.2432 \\
    R1-Distill-Qwen-14B \textit{(Zero-shot)} & 29.39 & 13.13 & 16.06 & 11.57 & 0.7920 \\
    R1-Distill-Qwen-14B \textit{(Few-shot)}  & 30.20 & 13.60 & 15.80 & 12.25 & 0.7548 \\
    Qwen2.5-72B \textit{(Zero-shot)} & 34.50 & 22.93 & 25.75 & 21.14 & 1.2180 \\
    Qwen2.5-72B \textit{(Few-shot)}  & 35.60 & 23.60 & 25.20 & 22.05 & 1.1112 \\
    \midrule
    Meta-Llama-3.1-8B \textit{(Zero-shot)} &  1.61 & 0.25 & 0.05 & 0.08 & 1.5004 \\
    Meta-Llama-3.1-8B \textit{(Few-shot)}  &  6.90 & 0.40 & 0.04 & 0.12 & 1.3817 \\
    Mistral-7B-v0.3 \textit{(Zero-shot)}   &  3.19 & 0.78 & 0.14 & 0.20 & 1.1479 \\
    Mistral-7B-v0.3 \textit{(Few-shot)}    & 10.30 & 1.05 & 0.12 & 0.24 & 1.0913 \\
    Qwen2.5-7B \textit{(Zero-shot)}        & 25.33 &  5.25 &  5.95 &  4.30 & 1.2411 \\
    Qwen2.5-7B \textit{(Few-shot)}         & 26.85 &  5.10 &  6.30 &  4.55 & 1.1371 \\
    \midrule
    $AlignSurvey_{\text{Mistral-7B-v0.3}}$     & 51.06 & 38.50 & 36.39 & 35.73 & 0.2097 \\
    $AlignSurvey_{\text{Meta-Llama-3.1-8B}}$   & 51.44 & 35.01 & 34.66 & 33.84 & 0.2137 \\
    $AlignSurvey_{\text{Qwen2.5-7B}}$          & 51.52 & 37.55 & 36.79 & 36.22 & 0.1849 \\
    \bottomrule
  \end{tabular}
  \caption{Task4 Structured Response Modeling Results on the \textsc{GSS}.}
  \label{tab:gss}
\end{table*}

Across all tasks, SurveyLM models demonstrate clear superiority over general-purpose LLMs, highlighting the importance and effectiveness of domain-specific fine-tuning using the AlignSurvey benchmark and dataset. The consistent and significant improvements emphasize the necessity of specialized model training to achieve robust and socially representative alignment in modeling human preferences and attitudes across diverse survey tasks.
\section{C. Extended Analysis}
\subsection{C.1 Ablation Studies}
\subsubsection{Task 1: Social Role Modeling.}
Table \ref{tab:task1_Ablation} provides detailed ablation results for Task 1, assessing the impact of foundational training and task-specific supervised fine-tuning (SFT).

Across all models (SurveyLM variants of Mistral-7B, Meta-Llama-3.1-8B, and Qwen2.5-7B), removing foundational training (“w/o Foundation”) results in moderate accuracy declines across attributes such as "Gender," "Income," and "Education," indicating that foundational training provides essential general knowledge beneficial for demographic inference. However, the omission of task-specific SFT (“w/o Task SFT”) causes severe performance degradation, notably in complex traits such as "Relationship" and "Expenses." For example, SurveyLM$_{\text{Qwen2.5-7B}}$ drops dramatically in predicting "Employment" from 52.43\% to 13.51\% without task-specific training.
This emphasizes that task-specific fine-tuning significantly contributes to model performance, particularly in accurately capturing nuanced social roles.

\begin{sidewaystable}
  \centering
  \small
  \begin{tabular}{l|ccccc|cccc|ccc}
    \toprule
    \multicolumn{1}{c|}{\textbf{Model}} &
    \multicolumn{5}{c|}{\textbf{Basic INFO}} &
    \multicolumn{4}{c|}{\textbf{Social INFO}} &
    \multicolumn{3}{c}{\textbf{Family INFO}} \\
    \cmidrule(lr){2-6}\cmidrule(lr){7-10}\cmidrule(lr){11-13}
     & \textbf{Gender} & \textbf{Age} & \textbf{Area} & \textbf{Edu.} & \textbf{Employ.} &
       \textbf{Stress} & \textbf{Home Own.} & \textbf{Veh. Own.} & \textbf{Income} &
       \textbf{Rel.} & \textbf{HH Size} & \textbf{Expenses} \\
    \midrule
    $SurveyLM_{\text{Mistral-7B-v0.3}}$ & 54.31\% & 45.87\% & 59.67\% & 37.90\% & 43.47\% & 35.44\% & 60.54\% & 60.71\% & 33.18\% & 85.88\% & 39.99\% & 29.28\% \\
    \textit{\ \ w/o Foundation}        & 52.92\% & 45.42\% & 58.75\% & 37.08\% & 41.67\% & 33.75\% & 59.17\% & 59.58\% & 32.08\% & 85.42\% & 39.17\% & 27.92\% \\
    \textit{\ \ w/o Task SFT}          & 37.96\% & 28.32\% & 23.53\% & 33.92\% & 37.51\% & 30.39\% & 49.45\% & 56.77\% & 29.61\% & 76.21\% & 35.95\% & 16.38\% \\
    \midrule
    $SurveyLM_{\text{Meta-Llama-3\_1-8B}}$ & 55.14\% & 47.03\% & 62.05\% & 41.02\% & 47.49\% & 44.02\% & 65.92\% & 65.01\% & 31.29\% & 80.06\% & 34.80\% & 32.78\% \\
    \textit{\ \ w/o Foundation}           & 53.75\% & 46.25\% & 60.42\% & 39.58\% & 46.67\% & 42.50\% & 64.58\% & 64.17\% & 30.42\% & 78.33\% & 33.75\% & 32.08\% \\
    \textit{\ \ w/o Task SFT}             & 39.64\% & 40.14\% & 57.69\% & 38.32\% & 44.66\% & 27.35\% & 60.94\% & 28.22\% & 18.90\% & 55.52\% & 28.99\% & 27.63\% \\
    \midrule
    $SurveyLM_{\text{Qwen2.5-7B}}$ & 67.81\% & 50.62\% & 60.58\% & 41.83\% & 52.43\% & 36.55\% & 68.25\% & 66.88\% & 40.42\% & 86.42\% & 34.04\% & 33.95\% \\
    \textit{\ \ w/o Foundation}      & 66.25\% & 49.58\% & 60.00\% & 41.25\% & 50.42\% & 35.00\% & 67.08\% & 65.42\% & 39.58\% & 84.17\% & 32.92\% & 32.50\% \\
    \textit{\ \ w/o Task SFT}        & 41.87\% & 37.83\% & 36.72\% & 33.83\% & 13.51\% & 18.40\% & 52.96\% & 55.29\% & 38.79\% & 41.42\% & 29.07\% & 20.31\% \\
    \bottomrule
  \end{tabular}
  \caption{Ablation results for Task~1 under full model, without foundational training, and without task‐specific SFT. Higher percentages indicate better performance.}
  \label{tab:task1_Ablation}
\end{sidewaystable}

\subsubsection{Task 2: Semi-Structured Interview Modeling.}
Table \ref{tab:task2_Ablation} illustrates ablation effects on interview response generation, focusing on "Naturalness," "Style Match," and "Consistency."

Interestingly, removing foundational training occasionally boosts "Naturalness," as seen with SurveyLM$_{\text{Mistral-7B}}$ (2.61 to 3.50), suggesting foundational training might introduce generality that can slightly constrain natural fluency. However, this removal negatively affects style alignment and consistency, especially for Qwen and Llama models. Task-specific fine-tuning remains indispensable, as its absence consistently decreases scores across all dimensions significantly, highlighting the critical role of task-specific adaptation in qualitative dialogue tasks.

\begin{table}[htp]
  \centering
  \small
  \begin{tabular}{lccc}
    \toprule
    \textbf{Model} & Naturalness & Style Match & Consistency \\
    \midrule
    $SurveyLM_\text{Mistral}$ & 2.61 & 2.90& 2.90\\
    \textit{ \ \ w/o Foundation} & 3.5 & 2.90 & 2.90  \\
    \textit{ \ \ w/o Task SFT} & 3.08 & 2.90 & 2.90 \\
    $SurveyLM_\text{Llama}$ & 3.77 & 3.00 & 2.94 \\
    \textit{ \ \ w/o Foundation} & 3.41 & 2.90 & 2.89 \\
    \textit{ \ \ w/o Task SFT} & 3.21 & 2.90 & 2.91 \\
    $SurveyLM_\text{Qwen}$ & 3.98 & 2.96 & 3.01\\
    \textit{ \ \ w/o Foundation} & 3.51 & 2.91 & 2.95\\
    \textit{ \ \ w/o Task SFT} & 3.15 & 2.52 & 2.69 \\
    \bottomrule
  \end{tabular}
  \caption{Ablation study for Task 2 Semi-Structured Interview Modeling. Scores (1 = worst, 5 = best) are reported for \textit{Naturalness}, \textit{Style Match}, and \textit{Consistency} on three SurveyLM backbones together with two variants of each: w/o Foundation (foundation pre-training removed) and w/o Task SFT (task-specific fine-tuning removed).}
  \label{tab:task2_Ablation}
\end{table}

\textbf{Task 3: Attitude Stance Modeling}
Table \ref{tab:task3_ablation} evaluates ablation results on various stance metrics, including Accuracy, Precision, Recall, F1, and Wasserstein Distance (WD).

The absence of foundational training moderately decreases overall performance. However, omitting task-specific fine-tuning causes drastic reductions across all models, notably in accuracy and F1 scores. For instance, SurveyLM$_{\text{Meta-Llama-3.1-8B}}$ experiences a sharp accuracy decline from 55.09\% to 36.25\%. WD increases notably, suggesting poor alignment with actual stance distributions. These results underscore the indispensable nature of task-specific training in effectively capturing nuanced attitudinal stances.

\begin{table*}[htp]
  \centering
  \small
  \begin{tabular}{lcccccccc}
    \toprule
    \textbf{Model} & ACC & PRE & RECALL & F1 & ROUGE & Jaccard & Cosine & WD \\
    \midrule
    $AlignSurvey_\text{Mistral}$ & 57.13\% & 54.44\% & 53.36\% & 53.85\% & 7.64 & 0.1308 & 0.0249 & 0.2974\\
    \textit{\ \ w/o Foundation} & 56.25\% & 53.57\% & 52.84\% & 53.01\% & 7.58 & 0.1292 & 0.0240 & 0.3026\\
    \textit{\ \ w/o Task SFT} & 43.75\% & 0.3871 & 0.3856 & 0.3349 & 6.82 & 0.1126 & 0.0101 & 0.3130\\
    \addlinespace
    $AlignSurvey_\text{Llama}$ & 55.09\% & 54.77\% & 51.02\% & 52.14\% & 7.86 & 0.1341 & 0.0189 & 0.4583\\
    \textit{\ \ w/o Foundation} & 54.37\% & 54.02\% & 50.14\% & 51.23\% & 7.75 & 0.1329 & 0.0182 & 0.4625\\
    \textit{\ \ w/o Task SFT} & 36.25\% & 0.3102 & 0.3262 & 0.2573 & 5.54 & 0.0940 & 0.0019 & 0.4840\\
    \addlinespace
    $AlignSurvey_\text{Qwen}$ & 56.83\% & 42.74\% & 41.49\% & 41.52\% & 7.69 & 0.1281 & 0.0172 & 0.3851\\
    \textit{\ \ w/o Foundation} & 56.25\% & 41.15\% & 40.91\% & 40.95\% & 7.65 & 0.1269 & 0.0165 & 0.3888\\
    \textit{\ \ w/o Task SFT} & 46.88\% & 61.09\% & 28.25\% & 21.94\% & 8.84 & 0.1110 & 0.0066 & 0.3970\\
    \bottomrule
  \end{tabular}
  \caption{Ablation study for Task 3: Alignment Verification. We report classification metrics plus text-similarity metrics ROUGE-L, Jaccard, cosine similarity of sentence embeddings (higher is better), and WD (lower is better).}
  \label{tab:task3_ablation}
\end{table*}

\textbf{Task 4: Structured Response Modeling}

Tables \ref{tab:oursurvey_ablation}, \ref{tab:chip_ablation}, and \ref{tab:gss_ablation} present ablation outcomes for structured survey response tasks across ASE, CHIP, and GSS datasets.

Removing foundational training consistently causes slight performance drops in accuracy and F1 scores across datasets, highlighting foundational training’s supporting role. Yet, removing task-specific SFT severely diminishes performance, indicated by substantial accuracy reductions and significantly elevated WD (e.g., SurveyLM$_{\text{Qwen}}$ on ASE goes from 55.47\% to 35.50\% accuracy and WD from 0.0541 to 2.002). These results reveal the critical necessity of task-specific fine-tuning in structured response modeling tasks to achieve accurate survey response alignment.

\begin{table*}[htp]
  \centering
  \small
  \begin{tabular}{lccccc}
    \toprule
    \textbf{Model} & ACC & PRE & RECALL & F1 & WD \\
    \midrule
    $AlignSurvey_\text{Mistral}$ & 54.59\% & 45.30\% & 41.01\% & 41.34\% & 0.0671\\
    \textit{\ \ w/o Foundation}  & 53.22\% & 44.12\% & 39.81\% & 39.57\% & 0.0765\\
    \textit{\ \ w/o Task SFT}    & 36.17\% & 23.94\% & 23.89\% & 21.13\% & 2.182\\[2pt]
    
    $AlignSurvey_\text{Llama}$   & 56.11\% & 44.97\% & 41.68\% & 41.47\% & 0.0675\\
    \textit{\ \ w/o Foundation}  & 54.59\% & 44.29\% & 39.12\% & 38.50\% & 0.0846\\
    \textit{\ \ w/o Task SFT}    & 34.33\% & 22.58\% & 19.70\% & 18.43\% & 2.098\\[2pt]
    
    $AlignSurvey_\text{Qwen}$    & 55.47\% & 44.23\% & 41.87\% & 41.79\% & 0.0541\\
    \textit{\ \ w/o Foundation}  & 54.86\% & 44.10\% & 39.55\% & 38.71\% & 0.8893\\
    \textit{\ \ w/o Task SFT}    & 35.50\% & 21.63\% & 18.62\% & 17.52\% & 2.002\\
    \bottomrule
  \end{tabular}
  \caption{Ablation results on \textbf{ASE}.}
  \label{tab:oursurvey_ablation}
\end{table*}

\begin{table*}[htp]
  \centering
  \small
  \begin{tabular}{lccccc}
    \toprule
    \textbf{Model} & ACC & PRE & RECALL & F1 & WD \\
    \midrule
    $AlignSurvey_\text{Mistral}$ & 65.56\% & 35.90\% & 38.49\% & 35.35\% & 0.0857\\
    \textit{\ \ w/o Foundation}  & 64.55\% & 33.87\% & 36.10\% & 32.54\% & 0.0912\\
    \textit{\ \ w/o Task SFT}    & 40.32\% & 34.31\% & 33.55\% & 29.72\% & 2.000\\[2pt]
    
    $AlignSurvey_\text{Llama}$   & 64.89\% & 32.91\% & 36.82\% & 32.99\% & 0.0845\\
    \textit{\ \ w/o Foundation}  & 64.55\% & 34.64\% & 36.76\% & 33.61\% & 0.0900\\
    \textit{\ \ w/o Task SFT}    & 40.03\% & 28.91\% & 29.45\% & 23.64\% & 1.984\\[2pt]
    
    $AlignSurvey_\text{Qwen}$    & 67.17\% & 36.48\% & 38.82\% & 34.94\% & 0.0686\\
    \textit{\ \ w/o Foundation}  & 64.07\% & 32.44\% & 36.30\% & 33.06\% & 0.0782\\
    \textit{\ \ w/o Task SFT}    & 30.87\% & 19.18\% & 19.16\% & 15.12\% & 1.992\\
    \bottomrule
  \end{tabular}
  \caption{Ablation results on \textbf{CHIP}.}
  \label{tab:chip_ablation}
\end{table*}

\begin{table*}[htp]
  \centering
  \small
  \begin{tabular}{lccccc}
    \toprule
    \textbf{Model} & ACC & PRE & RECALL & F1 & WD \\
    \midrule
    $AlignSurvey_\text{Mistral}$ & 51.06\% & 38.50\% & 36.39\% & 35.73\% & 0.2097\\
    \textit{\ \ w/o Foundation}  & 50.34\% & 34.43\% & 34.88\% & 33.73\% & 0.3205\\
    \textit{\ \ w/o Task SFT}    & 50.97\% & 37.10\% & 35.34\% & 34.56\% & 2.008\\[2pt]
    
    $AlignSurvey_\text{Llama}$   & 51.44\% & 35.01\% & 34.66\% & 33.84\% & 0.2137\\
    \textit{\ \ w/o Foundation}  & 51.48\% & 35.02\% & 34.45\% & 33.37\% & 0.2209\\
    \textit{\ \ w/o Task SFT}    & 50.34\% & 34.60\% & 34.13\% & 33.05\% & 1.977\\[2pt]
    
    $AlignSurvey_\text{Qwen}$    & 51.52\% & 37.55\% & 36.79\% & 36.22\% & 0.1849\\
    \textit{\ \ w/o Foundation}  & 51.63\% & 37.94\% & 36.13\% & 35.46\% & 0.2118\\
    \textit{\ \ w/o Task SFT}    & 50.47\% & 35.65\% & 34.59\% & 33.67\% & 1.992\\
    \bottomrule
  \end{tabular}
  \caption{Ablation results on \textbf{GSS}.}
  \label{tab:gss_ablation}
\end{table*}

\subsection{C.2 Analysis by Demographic Sub-groups}

\begin{figure*}[htp]
    \centering
    \includegraphics[width=\textwidth]{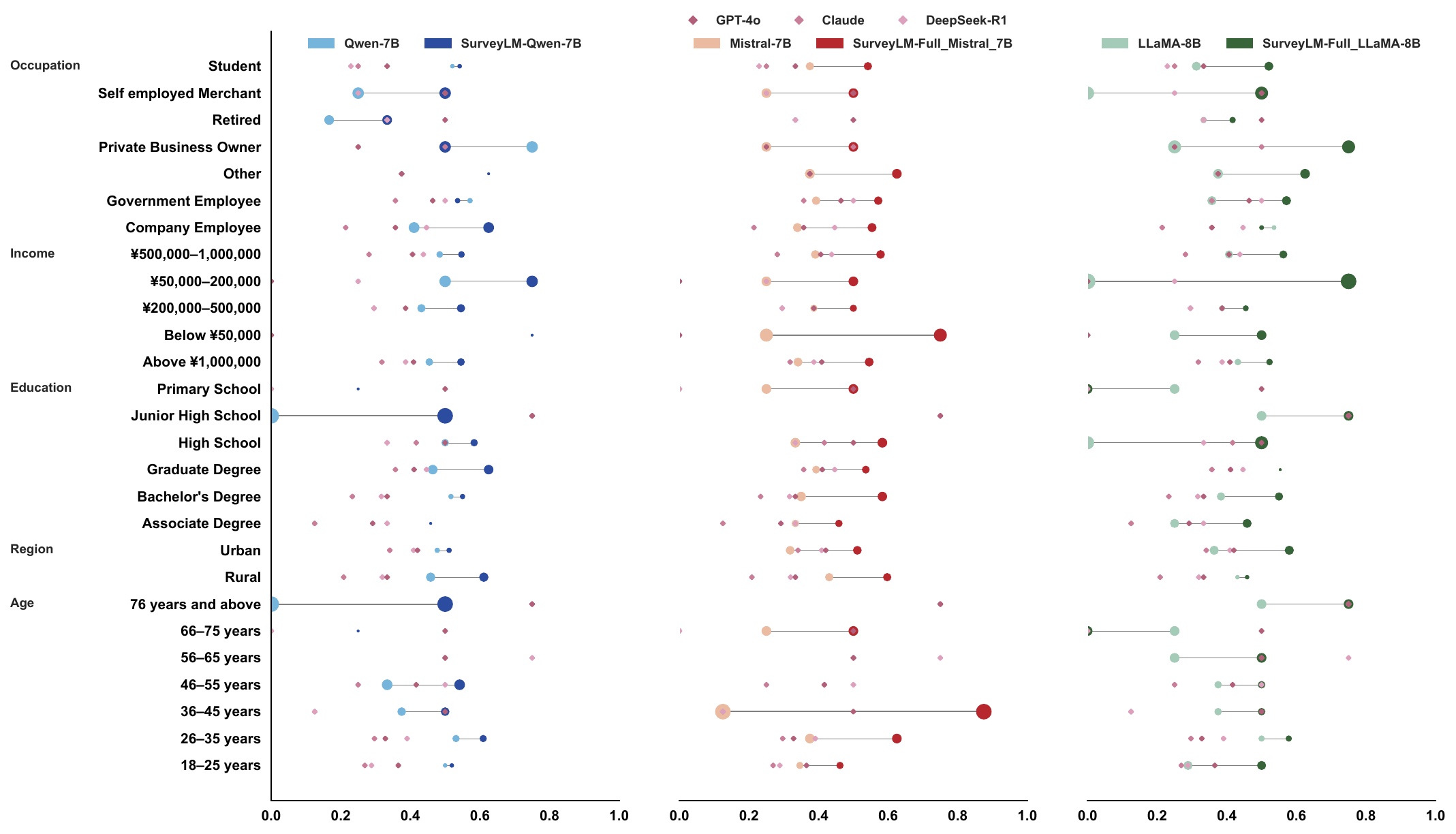}
    \caption{Multi-Demographic Accuracy Comparison in Task3. Circle size represents the accuracy gap between Qwen-7B, Mistral-7B and LLaMA-8B, with their SurveyLM. Gains are especially pronounced for under-represented groups (rural residents, aged 76+, self-employed, low- and middle- income groups).}
    \label{task3_ana}
\end{figure*}

Figure \ref{task3_ana} provides demographic subgroup analysis on attitude stance modeling.
SurveyLM models outperform baseline general-purpose models across demographic groups, notably closing performance gaps for marginalized populations such as elderly individuals (76+ years), rural residents, and low-income groups (below ¥50,000). SurveyLM$_{\text{Qwen2.5-7B}}$ demonstrates substantial accuracy improvements across education and occupation categories, notably benefiting groups with primary education and self-employed merchants. General models (GPT-4o, Claude, DeepSeek-R1) show markedly lower performance for these underrepresented groups, underscoring SurveyLM’s enhanced capability in addressing demographic disparities.

\begin{figure*}[htp]
    \centering
    \includegraphics[width=\textwidth]{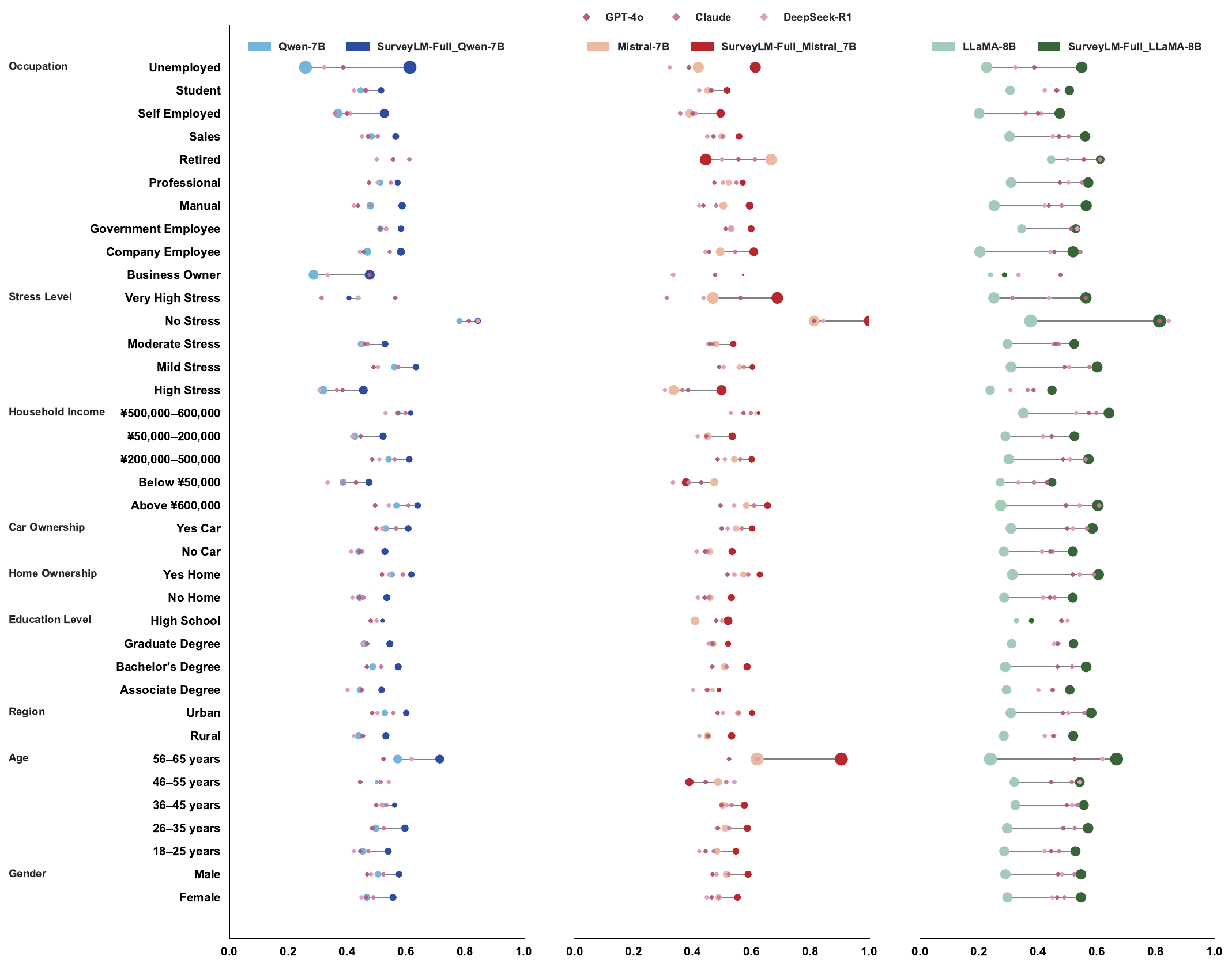}
    \caption{Multi-Demographic Accuracy Comparison in Task4. Circle size represents the accuracy gap between Qwen-7B, Mistral-7B and LLaMA-8B, with their SurveyLM.}
    \label{task4ana}
\end{figure*}

Figure \ref{task4ana} displays demographic subgroup performance analysis for structured response tasks.
SurveyLM models consistently demonstrate higher accuracy across all demographic segments compared to baseline models. Particularly significant improvements are observed among groups with lower education levels (High School), manual workers, unemployed, and students, typically underrepresented in surveys. For instance, SurveyLM$_{\text{Mistral-7B}}$ dramatically improves performance in predicting responses from manual workers and those experiencing high stress levels. Contrarily, GPT-4o and Claude models consistently underperform in accurately capturing responses from less advantaged demographic groups.

These demographic subgroup analyses confirm SurveyLM models' superior alignment with diverse populations, effectively reducing response biases and improving representation accuracy across varied social and economic backgrounds.

Extended analyses through ablation and demographic subgroups robustly confirm the essential role of task-specific fine-tuning and foundational training for achieving high-quality, fair, and demographically representative outcomes. SurveyLM models demonstrate remarkable improvements across demographic diversity, validating the significance of targeted model adaptation in enhancing both performance and fairness in social survey modeling.
\section{D. Task Prompt Templates}
\subsection{D.1 Task 1 Social Role Modeling}

The prompt template provided in Table \ref{tab:task1_prompt} for Task 1 focuses on social role modeling by leveraging dialogue-based interactions. The task simulates an analytical scenario where the model, acting as an interview analyst, predicts demographic information (in this case, age) based on conversational cues provided within an interview.

The instruction explicitly defines the role of the model as an analyst tasked with interpreting survey dialogue to predict demographic attributes. The dialogue excerpt provided serves as a realistic representation of how respondents typically articulate their understanding and perception of social issues. The prompt clearly defines a query, which in this scenario is the respondent's age, providing discrete age ranges as potential outcomes.

\begin{table*}[htp]
  \centering
  \begin{tabular}{p{\linewidth}}
    \toprule
    \makecell[{{p{\linewidth}}}]{%
      \textbf{INSTRUCTION:} You are an interview questionnaire analyst, analyzing a survey interview text about people's perceived understanding of common prosperity.\\
      \\
      \textbf{==Interview Dialogue==}\\
      Interviewer: Okay, then, in your mind, what should common prosperity look like?\\
      Interviewee: I believe common prosperity has two components. First, it's universal prosperity, meaning it encompasses all people, not just a few. Second, I believe common prosperity is comprehensive. Beyond a sense of material prosperity, it also requires spiritual enrichment.%
    }\\
    \midrule
    \textbf{QUERY:} Based on the conversation, please predict the respondent's \emph{age}.\\
    \\
    Optional values: 18–25, 26–35, 36–45, 46–55, 56–65, 66–75, 76 and older.\\
    \midrule
    \textbf{ANSWER:} 26–35 years old\\
    \bottomrule
  \end{tabular}
  \caption{The Prompt Temple of Task1.}
  \label{tab:task1_prompt}
\end{table*}

\subsection{D.2 Task 2 Semi-Structured Interview Modeling}
Task 2, illustrated in Table \ref{tab:task2_prompt}, represents a semi-structured interview modeling task. Here, the prompt situates the model as a survey respondent, characterized by specific demographic and socioeconomic attributes provided in detail within the background information section. The model is expected to produce responses consistent with this detailed profile, thereby assessing its ability to simulate realistic, context-aware human-like conversational behaviors.

The instructions delineate the scenario clearly, including previous conversational context, which the model must consider to ensure coherence and realism. The current question "How would you put it?" prompts the model to generate a response consistent with the character's socioeconomic status, education level, living situation, and employment status, reflecting nuanced awareness of subtle social distinctions.

\begin{table*}[htp]
  \centering
  \begin{tabular}{p{\linewidth}}
    \toprule
    \makecell[{{p{\linewidth}}}]{
      \textbf{INSTRUCTION:}
      You are a respondent conducting a survey on the public's perceived understanding of common prosperity.\\
      \\
      \textbf{==Your Background==}\\
      Gender: Male\\
      Age: 26--35\\
      Place of Household Registration: Taizhou, Zhejiang Province\\
      Permanent Residence: Hangzhou, Zhejiang Province\\
      Household Registration Status: Rural\\
      Highest Education Level: Graduate or above\\
      Number of Family Members: 3\\
      Family Membership: Children, Grandparents\\
      Property Status: Yes\\
      Vehicle Status: No\\
      Annual Household Income: Over 1 million\\
      Major Monthly Household Expenditure Items: Daily Living Expenses (food, clothing, etc.); Education Expenses (tuition, supplementary materials, etc.); Healthcare Expenses (medical insurance, medicine, medical treatment, etc.)\\
      Life Stress Level: Mild Stress -- occasionally feel stressed, but it does not affect life\\
      Occupation Type: Corporate Employee\\
      \\
      \textbf{==Previous Conversation==}\\
      Question: Zhejiang began developing the Common Prosperity Demonstration Zone in 2021. Do you think there have been any changes compared to the past, especially before the establishment of the Common Prosperity Demonstration Zone?\\
      Answer: I personally think the difference is not significant.\\
      \\
      \textbf{==Current Question==}\\
      How would you put it?
    }\\
    \midrule
    \textbf{QUERY:} Please enter your answer: \\
    \midrule
    \textbf{ANSWER:} Maybe it is related to the industry I work in. I work in the Internet industry. Relatively speaking, the overall salary of the industry and the company has not changed much in the past two years.\\
    \bottomrule
  \end{tabular}
  \caption{The Prompt Temple of Task2.}
  \label{tab:task2_prompt}
\end{table*}

\subsection{D.3 Task 3 Attitude Stance Modeling}
The prompts in Tables \ref{tab:task3.1_prompt} and \ref{tab:task3.2_prompt} outline two critical subtasks within attitude stance modeling. Task 3.1 requires the model to analytically determine and rationalize the respondent's stance towards public services based on demographic data and implied context. The instruction specifies an output format, demanding structured, interpretable reasoning that explicitly justifies the model’s stance prediction in JSON format.

Task 3.2 is structured quantitatively, requiring the model to predict response distribution across positive, neutral, and negative attitudes toward common prosperity among a specified demographic subgroup (26-35 years). The prompt explicitly demands a percentage-based distribution output, emphasizing the model’s capability to aggregate nuanced attitude information into accurate, representative quantitative summaries.

\begin{table*}[htp]
  \centering
  \begin{tabular}{p{\linewidth}}
    \toprule
    \makecell[{{p{\linewidth}}}]{
      \textbf{INSTRUCTION:}
      You are a sociology researcher analyzing interview data from a respondent:\\
      \\
      \textbf{==Your Background==}\\
      Gender: Male\\
      Age: 26--35\\
      Place of Household Registration: Taizhou, Zhejiang Province\\
      Permanent Residence: Hangzhou, Zhejiang Province\\
      Household Registration Status: Rural\\
      Highest Education Level: Graduate or above\\
      Number of Family Members: 3\\
      Family Membership: Children, Grandparents\\
      Property Status: Yes\\
      Vehicle Status: No\\
      Annual Household Income: Over 1 million\\
      Major Monthly Household Expenditure Items: Daily Living Expenses (food, clothing, etc.); Education Expenses (tuition, supplementary materials, etc.); Healthcare Expenses (medical insurance, medicine, medical treatment, etc.)\\
      Life Stress Level: Mild Stress -- occasionally feel stressed, but it does not affect life\\
      Occupation Type: Corporate Employee\\
    }\\
    \midrule
    \textbf{QUERY:} Please determine the respondent's evaluation of public services and explain your reasoning. \\
The output format is JSON and contains two fields: "predict" and "reason." The predict field is selected from "positive", "negative" or "neutral". \\
    \midrule
    \textbf{ANSWER:} \{"predict": "Positive", "reason": "Respondents have lived in county towns for a long time and may have been influenced by the renovation of older residential areas and government public service improvement projects. Therefore, they have a relatively positive attitude towards public services, believing them to have "improved somewhat." Regarding education and healthcare, respondents expressed limited understanding. This may be due to the health of their elderly family members and their current childlessness, which may result in relatively low demand for education and healthcare, and limited experience or feelings about them."\}\\
    \bottomrule
  \end{tabular}
  \caption{The Prompt Temple of Task3.1.}
  \label{tab:task3.1_prompt}
\end{table*}

\begin{table*}[htp]
  \centering
  \begin{tabular}{p{\linewidth}}
    \toprule
    \makecell[{{p{\linewidth}}}]{
      \textbf{INSTRUCTION:}
      Please provide the distribution of responses to the following question for people aged 26-35:\\
      \\
      What is your attitude towards common prosperity? \\
      1. Positive \\
      2. Neutral \\
      3. Negative\\
    }\\
    \midrule
    \textbf{QUERY:} Please provide the distribution of responses (in percentages), for example, \\
    A.xx: xx\% \\
    B.xx: xx\% \\
    C.xx: xx\% \\
    \midrule
    \textbf{ANSWER:} \{"Positive": "40\%", "Neutral": "46\%", "Negative": "13\%"\} \\
    \bottomrule
  \end{tabular}
  \caption{The Prompt Temple of Task3.2.}
  \label{tab:task3.2_prompt}
\end{table*}

\subsection{D.4 Task 4 Structured Response Modeling}

The structured response modeling task, described in Tables \ref{tab:task4.1_prompt} and \ref{tab:task4.2_prompt}, focuses on both individual and aggregate structured responses to survey questions. Task 4.1 simulates a realistic scenario wherein the model represents a respondent with explicitly provided demographic and socioeconomic contexts. This explicit profile requires the model to generate responses that accurately reflect likely perspectives based on specified individual attributes, such as place of residence, income, family structure, and occupation.

Task 4.2 extends structured response modeling by requiring demographic-specific distribution predictions for community engagement attitudes among office-working females with moderate household incomes. This task demands sophisticated aggregation capabilities, as the model must synthesize demographic knowledge to generate believable statistical distributions. The explicitly percentage-based format tests both demographic sensitivity and the model’s ability to accurately capture population-level variability and representativeness in responses.

\begin{table*}[htp]
  \centering
  \begin{tabular}{p{\linewidth}}
    \toprule
    \makecell[{{p{\linewidth}}}]{
      \textbf{INSTRUCTION:}
      Survey time: 2025. You need to simulate a real survey taker to answer the following questions.\\
      \\
      \textbf{==Your Background==}\\
      Gender: Male\\
      Age: 26--35\\
      Place of Household Registration: Taizhou, Zhejiang Province\\
      Permanent Residence: Hangzhou, Zhejiang Province\\
      Household Registration Status: Rural\\
      Highest Education Level: Graduate or above\\
      Number of Family Members: 3\\
      Family Membership: Children, Grandparents\\
      Property Status: Yes\\
      Vehicle Status: No\\
      Annual Household Income: Over 1 million\\
      Major Monthly Household Expenditure Items: Daily Living Expenses (food, clothing, etc.); Education Expenses (tuition, supplementary materials, etc.); Healthcare Expenses (medical insurance, medicine, medical treatment, etc.)\\
      Life Stress Level: Mild Stress -- occasionally feel stressed, but it does not affect life\\
      Occupation Type: Corporate Employee\\
    }\\
    \midrule
    \textbf{QUERY:} To what extent do you believe the economy of your county (city, district) is developing rapidly?  \\
    Options: \\
    A. Neutral \\
    B. Agree \\
    C. Disagree \\
    Please enter your answer directly. \\
    \midrule
    \textbf{ANSWER:} B. Agree\\
    \bottomrule
  \end{tabular}
  \caption{The Prompt Temple of Task4.1.}
  \label{tab:task4.1_prompt}
\end{table*}

\begin{table*}[htp]
  \centering
  \begin{tabular}{p{\linewidth}}
    \toprule
    \makecell[{{p{\linewidth}}}]{
      \textbf{INSTRUCTION:}
      Please provide the distribution of the following groups of people across the questionnaire options below.\\
      \\
      \textbf{==Population Characteristics==}\\
      Gender: Female\\
      Occupation: Office Worker\\
      Annual Household Income: 110,000-200,000 RMB}\\
    \midrule
    \textbf{QUERY:} To what extent do you believe you can participate in community or village affairs?  \\
    Options: \\
    A. Agree \\
    B. Neutral \\
    C. Disagree \\
    Please provide the distribution of answers directly (in percentage form), for example: \\
    A. xx\% \\
    B. xx\% \\
    C. xx\% \\
    \midrule
    \textbf{ANSWER:} \\
    Neutral: 47\% \\
    Disagree: 16\% \\
    Agree: 36\% \\
    \bottomrule
  \end{tabular}
  \caption{The Prompt Temple of Task4.2.}
  \label{tab:task4.2_prompt}
\end{table*}

By encompassing diverse survey response formats—from open-ended dialogues to structured quantitative responses—the provided templates offer versatile benchmarks essential for developing and refining socially aligned language modeling technologies. Consequently, they serve as foundational tools for systematically advancing model interpretability, accuracy, realism, and social representativeness in human-oriented AI applications.
\section{E. Automatic LLM-based Evaluators}

\subsection{E.1 Evaluator Prompt Template}
The prompt used for the automatic evaluators is shown in Table~\ref{tab:evaluator}.

\subsection{E.2 Scoring Dimensions and Rating Criteria}
The evaluator scores each mock answer along three dimensions: \textbf{Naturalness}, \textbf{Style Match}, and \textbf{Consistency}.  Each dimension is rated on a 1 – 5 ordinal scale.

\begin{table*}[htp]
  \centering
  \begin{tabular}{p{\linewidth}}
    \toprule
    \makecell[{{p{\linewidth}}}]{
    \textbf{INSTRUCTION:}
    Mock Interview Response Evaluation Criteria\\
    \\
    \textbf{Scoring Dimensions}\\
    1. Naturalness: Evaluate whether the interview response is expressed in natural language, connects well with the questions, and reflects the tone and rhythm of real-life conversation. Avoid templated or bulleted responses. **If the response is too structured, the score for this item should be appropriately reduced. \\
    2. Style Match: Evaluate whether the response is consistent with the standard response, including language style, thinking style, and focus. \\
    3. Consistency: Determine whether the response accurately expresses the core points related to the question and covers the main positions or logic reflected in the standard answer. **If the response expresses judgments not reflected in the standard answer, the score for this item should be appropriately reduced.\\
    \\
    \textbf{Rating Criteria} \\
    Score the mock answers based on the above dimensions, using a 1-5 scale for each dimension. \\
    - 5: Fully meets the evaluation criteria, with accurate, natural, and complete expression. \\
    - 4: Mostly meets the criteria, with minor deficiencies or deviations. \\
    - 3: Partially meets the criteria, with significant problems or omissions. \\
    - 2: Only partially meets the criteria, with numerous problems that hinder comprehension. \\
    - 1: Completely fails to meet the criteria, with disorganized structure or significant deviation from the topic. \\
    \\
    \textbf{==Your Background==}\\
    Gender: Male\\
    Age: 26--35\\
    Place of Household Registration: Taizhou, Zhejiang Province\\
    Permanent Residence: Hangzhou, Zhejiang Province\\
    Household Registration Status: Rural\\
    Highest Education Level: Graduate or above\\
    Number of Family Members: 3\\
    Family Membership: Children, Grandparents\\
    Property Status: Yes\\
    Vehicle Status: No\\
    Annual Household Income: Over 1 million\\
    Major Monthly Household Expenditure Items: Daily Living Expenses (food, clothing, etc.); Education Expenses (tuition, supplementary materials, etc.); Healthcare Expenses (medical insurance, medicine, medical treatment, etc.)\\
    Life Stress Level: Mild Stress -- occasionally feel stressed, but it does not affect life\\
    Occupation Type: Corporate Employee\\
    \\
    \textbf{==Previous Conversation==}\\
    Question: Zhejiang began developing the Common Prosperity Demonstration Zone in 2021. Do you think there have been any changes compared to the past, especially before the establishment of the Common Prosperity Demonstration Zone?\\
    Answer: I personally think the difference is not significant.\\
    \\
    \textbf{==Current Question==}\\
    How would you put it?
    \textbf{Model Answer} \\
    Maybe it is related to the industry I work in. I work in the Internet industry. Relatively speaking, the overall salary of the industry and the company has not changed much in the past two years.\\
    \\
    \textbf{Mock Answer} \\
    Being part of the Internet industry might explain it. Over the past two years, compensation for both the sector as a whole and my employer in particular has remained largely unchanged.
    } \\
    \midrule
    \textbf{QUERY:} Output Format 
    \{\{ "Naturalness": \{\{ "Rating": "X", "Comment": "..." \}\}, "Style Match": \{\{ "Rating": "Y", "Comment": "..." \}\}, "Consistency": \{\{ "Rating": "Z", "Comment": "..." \}\} \}\} \\
    \midrule
    \textbf{ANSWER:} 
    \{"Naturalness": \{ "Rating": "4", "Comment": "The simulated answers were expressed in natural language and connected to the questions reasonably, but lacked specific examples, making the conversation a bit vague.\},
    "Style Match": \{ "Rating": "3", "Comment": "The focus of the answer deviated from the specific examples in the standard answer, while the language style remained consistent but slightly vague.\}, 
    "Consistency": \{"Rating": "3", "Comment": "The specific examples in the standard answer were inconsistent, failing to accurately express the specific perception of public service use and lacking specific examples.\}\\
    \bottomrule
  \end{tabular}
  \caption{The Prompt Temple of Evaluator in Task2.}
  \label{tab:evaluator}
\end{table*}

\subsection{E.3 Alignment Evaluation}

\begin{table}[htbp]
  \centering
  \begin{tabular}{lcc}
    \toprule
    \textbf{Statistic} & \textbf{Human–Human} & \textbf{Human–LLM} \\
    \midrule
    Fleiss’ $\kappa$ & 0.68 & — \\
    Krippendorff’s $\alpha$ & 0.72 & — \\
    Pearson $r$ & — & 0.74 \\
    Spearman $\rho$ & — & 0.73 \\
    \bottomrule
  \end{tabular}
  \caption{Inter-annotator reliability and human–model alignment statistics. Values above 0.60 (reliability) and 0.70 (correlation) are generally considered acceptable.}
  \label{tab:agreement}
\end{table}

To verify the reliability of the automatic LLM scores, we first collected pairwise comparative judgments from domain experts.  
For each prompt, three experts compared two model responses on overall preference.
The consistency of our annotation rubric was assessed with both  
\textbf{Fleiss’~$\kappa$} and \textbf{Krippendorff’s~$\alpha$}.  Both statistics slightly exceed the commonly accepted 0.60 threshold, indicating \emph{moderate} but acceptable inter-rater agreement.

We then examined the alignment between aggregated human scores and those produced by an automatic LLM-based evaluator (Table \ref{tab:agreement}).  Pearson and Spearman correlations in the higher than 0.70 show that the evaluator captures human preferences reasonably well while providing orders-of-magnitude higher throughput.
\section{F. Annotation Reliability}

To operationalise Task 3 Attitude Stance Modeling, we first created a concise codebook that maps every dialogue segment to one of three mutually–exclusive stance labels: \textit{positive (1)}, \textit{neutral (2)}, or \textit{negative (3)}.  Three graduate-level social-science researchers were recruited.  Throughout production coding, annotators worked independently and blind to one another.

Every dialogue segment (N = 644) received three independent labels.  
Disagreements triggered a two-step adjudication:

\begin{itemize}
  \item \textbf{Automatic majority rule}: if two of three raters agreed, their label became the provisional gold.  
  \item \textbf{Expert arbitration}: segments with a three–way split were forwarded to a senior domain expert, who reviewed the raw transcript together with the annotators’ written rationales and issued a final decision.
\end{itemize}

We computed three complementary reliability statistics:

\begin{table}[t]
  \centering
  \begin{tabular}{llcc}
    \toprule
    Statistic & Rater(s) Compared & $\kappa$ / $\alpha$ & $P_o$ \\
    \midrule
    \multirow{3}{*}{Cohen’s $\kappa$} 
      & people1 vs.\ people2 & 0.63 & 0.793 \\
      & people1 vs.\ people3 & 0.71 & 0.852 \\
      & people2 vs.\ people3 & 0.66 & 0.811 \\
    \midrule
    Fleiss’ $\kappa$        & all three & 0.67 & 0.822 \\
    Krippendorff’s $\alpha$ & all three & 0.74 & -- \\
    \bottomrule
  \end{tabular}
    \caption{Inter-coder agreement statistics for Task 3.}
  \label{tab:reliability}
\end{table}

Taken together, these scores show that annotators converged well above chance and satisfy the minimum thresholds ($\geq$ 0.60 for $\kappa$, $\geq$ 0.67 for $\alpha$) recommended for corpus construction in empirical social-science research.

Although the stance scale is deliberately fine-grained, moderate–to–substantial agreement was achieved without prolonged training, suggesting that the codebook captures readily observable cues in ordinary language.  The residual disagreements mainly stem from (i) sarcastic or ambivalent statements and (ii) domain-specific jargon.  Future work can further boost consistency by providing additional sarcasm exemplars and a mini-glossary for specialised terms.
\section{G. Licensing, Ethics \& Societal Impact}
\subsection{G.1 Data Licenses and Usage Rights}

All datasets used in the AlignSurvey project were sourced from publicly accessible platforms or officially released academic resources, and processed solely for non-commercial research purposes in accordance with their respective usage terms and academic licensing policies.

\textbf{Video and Interview Text Data:}  
The data were collected from publicly accessible platforms and oral history materials, restricted to content openly available without access limitations. All processing was conducted for academic use only. The data construction process involved transcription and anonymization of spoken content. No original video, audio, or full dialogue content is stored, retained, or distributed. All downstream resources are derived exclusively from transcribed text and expert annotations. Prior peer-reviewed research has demonstrated the value and academic acceptability of using content from public video platforms for large-scale benchmarking and social analysis \cite{real2017youtube,loh2022youtube,albadi2022deradicalizing}. Our approach is consistent with this line of work and adheres to public content usage norms and academic best practices.



\textbf{Social Survey Data:}
We incorporate publicly released, academically licensed survey datasets including the American Trends Panel (ATP), European Social Survey (ESS), General Social Survey (GSS), Chinese Social Survey (CSS), Chinese General Social Survey (CGSS), and China Household Income Project (CHIP). All datasets are used in strict accordance with their official licensing terms and intended for non-commercial academic research only. Personally identifiable information was thoroughly removed following established anonymization procedures to ensure participant privacy.



\textbf{Data Distribution and Public Release:}
AlignSurvey-related data, tools, and model parameters are publicly released via platforms such as GitHub and HuggingFace under the Creative Commons Attribution–NonCommercial–NoDerivatives 4.0 International License (CC BY-NC-ND 4.0). All resources are intended strictly for non-commercial academic research. Users must strictly adhere to both the open-source license terms and the licensing agreements of any original data sources included in the benchmark.


\subsection{G.2 Model Release and License Compliance} 
The SurveyLM model family released by AlignSurvey is based on open-source large language models (LLMs), refined through a two-stage fine-tuning process. Initially, models are trained on a diverse social-context corpus, encompassing multilingual dialogues and structured survey texts to develop broad social-contextual awareness. Subsequently, these models undergo task-specific supervised fine-tuning with comprehensive datasets tailored to each research task. All model checkpoints are released strictly for non-commercial academic research purposes.


\textit{Mistral 7B Model:}
Released under the Apache 2.0 License, which allows broad usage—including modification and redistribution—provided that proper attribution and copyright notices are retained.

\textit{Meta Llama 3.1 Model:}
Released under Meta’s COMMUNITY LICENSE AGREEMENT. This license permits non-commercial research use and derivative works, while prohibiting commercial deployment without separate authorization.

\textit{Qwen-7B Model:}
Also released under the Apache 2.0 License, permitting academic and commercial use with proper attribution, in accordance with the license terms published by Qwen's maintainers.

While the base models support varying degrees of commercial use, all SurveyLM models released as part of the AlignSurvey benchmark are explicitly scoped for non-commercial academic use only. Distributed model weights include clear notices reinforcing original license terms, derivative usage conditions, and the research-only scope of redistribution.

\subsection{G.3 Ethical Compliance and Privacy Protection}

AlignSurvey adheres to rigorous ethical and privacy standards throughout data curation and model training:

\textbf{Privacy by Design:}  
All interview and survey content underwent strict de-identification and removal of any potentially sensitive information. Dataset construction avoided collecting or storing any original media content (e.g., video, audio, or images).

\textbf{Minimal Use and Responsible Processing:}  
For publicly sourced materials, only transcribed and anonymized interview text was retained. No raw media content, full transcripts, or personally revealing information is included or shared. This aligns with academic norms and ensures the minimal-use principle is upheld.

\textbf{Academic Usage Boundaries:}  
All released resources are restricted to non-commercial academic research. They are not intended for use in clinical, legal, or other high-stakes decision-making scenarios without appropriate domain expertise and human oversight. 

\textbf{Inclusivity and Fairness:}  
Throughout dataset construction and model evaluation, efforts were made to ensure broad demographic representation and minimize systematic bias. Particular care was given to underrepresented or marginalized groups, in both sampling and evaluation criteria.






\subsection{G.4 Responsible Use and Ethical Safeguards}

To support responsible innovation and uphold high standards of academic integrity, we implement the following safeguards:

\textbf{Content Quality Assurance:}  
Prior to release, model outputs were reviewed through a combination of automated tools and human-in-the-loop checks to ensure thematic relevance and discursive appropriateness. Users are encouraged to verify fitness for their specific research scenarios in accordance with institutional practices.


\textbf{Usage Guidelines:}  
The project provides clear documentation regarding usage scope, licensing, and intended applications. Users are expected to follow institutional research guidelines and uphold responsible academic practices.

These safeguards are designed to foster constructive use of large language models in computational social science while promoting transparency, inclusivity, and scholarly rigor.

\end{document}